\documentclass[conference]{IEEEtran}
\IEEEoverridecommandlockouts
\usepackage{cite}
\usepackage{amsmath,amssymb,amsfonts}
\usepackage{algorithmic}
\usepackage{textcomp}
\usepackage{xcolor}
\usepackage{url}
\usepackage{graphicx}
\usepackage[subrefformat=parens]{subcaption}
\usepackage{multirow}
\usepackage{comment}
\usepackage{color}

\def\BibTeX{{\rm B\kern-.05em{\sc i\kern-.025em b}\kern-.08em
    T\kern-.1667em\lower.7ex\hbox{E}\kern-.125emX}}

\begin{document}

\title{Visual Explanation using Attention Mechanism in Actor-Critic-based Deep Reinforcement Learning}

\author{
\IEEEauthorblockN{
Hidenori Itaya, Tsubasa Hirakawa, Takayoshi Yamashita, Hironobu Fujiyoshi,
}
\IEEEauthorblockA{Chubu University\\
1200 Matsumotocho, Kasugai, Aichi, Japan \\
\texttt{\{itaya, hirakawa\}@mprg.cs.chubu.ac.jp, \{takayoshi, fujiyoshi\}@isc.chubu.ac.jp}
}
\\
\IEEEauthorblockN{
Komei Sugiura
}
\IEEEauthorblockA{Keio University \\
3-14-1 Hiyoshi, Kohoku, Yokohama, Kanagawa, Japan \\
\texttt{komei.sugiura@keio.jp}
}
}

\maketitle

\begin{abstract}
Deep reinforcement learning (DRL) has great potential for acquiring the optimal action in complex environments such as games and robot control.
However, it is difficult to analyze the decision-making of the agent, i.e., the reasons it selects the action acquired by learning.
In this work, we propose Mask-Attention A3C (Mask A3C), which introduces an attention mechanism into Asynchronous Advantage Actor-Critic (A3C),  which is an actor-critic-based DRL method, and can analyze the decision-making of an agent in DRL.
A3C consists of a feature extractor that extracts features from an image, a policy branch that outputs the policy, and a value branch that outputs the state value.
In this method, we focus on the policy and value branches and introduce an attention mechanism into them.
The attention mechanism applies a mask processing to the feature maps of each branch using mask-attention that expresses the judgment reason for the policy and state value with a heat map.
We visualized mask-attention maps for games on the Atari 2600 and found we could easily analyze the reasons behind an agent's decision-making in various game tasks.
Furthermore, experimental results showed that the agent could achieve a higher performance by introducing the attention mechanism.
\end{abstract}


\section{Introduction}
Reinforcement learning (RL) problems seek optimal actions to maximize cumulative rewards.
Unlike supervised learning problems, RL problems collect training data by exploring the environment. 
Therefore, RL has achieved high performance in specific tasks (e.g., controlling autonomous systems \cite{rl-robot:kober2013,rl-robot:gu2017,rl-robot:rajeswaran2017} and video games \cite{rl-game:tessler2017,rl-game:justesen2017,rl-game:shao2019}) in which it is difficult to create training data.
In Go, AlphaGo has defeated a professional Go player \cite{alphago}.
In 2015, the deep Q-network (DQN), a method that combines Q-learning \cite{ql} and deep neural network (DNN), achieved a score higher than human players on the Atari 2600 \cite{dqn-nature}.
Since the advent of DQN, deep RL (DRL), a method that combines deep learning and RL, has become mainstream, and it is now possible to solve problems featuring a huge number of states, such as images.\par
In general, deep learning can solve complex tasks by training using a large number of network parameters.
However, it is difficult to understand the reasoning behind the decision-making of the trained network because the number of network parameters used to make the decision is enormous.
This problem occurs in DRL as well.
The reason for judging the acquired action is unclear, since agents collect training data by searching the environment and the calculation inside the network is complicated.
Therefore, in order to prove that the trained network is sufficiently reliable, it is important to analyze the reason for the judgment of the action that it outputs.\par
One approach to interpreting the decision-making of a network, visual explanation, has been studied in the field of computer vision \cite{cam,grad-cam,abn}. 
Visual explanations analyze the factors of the network output by using an attention map that highlights the important regions in an input image. 
Visual explanation methods have also been applied to DRL models to help with understanding the decision-making of an agent \cite{rlatt:darqn,rlatt:greydanus}. 
These methods can be categorized into two approaches: bottom-up and top-down.
Bottom-up visual explanations compute attention maps by using the gradient information of a network. 
Because the bottom-up approach does not need to re-train a network, it can be applied to any trained network and is commonly used in computer vision and DRL.
The attention maps obtained by the bottom-up approach are based on the input data and response values calculated from each layer.
The bottom-up approach highlights local textural context.
Top-down visual explanations generate attention maps by using the response values in a network.
In contrast to the bottom-up approach, the attention maps of the top-down approach are output for the current network output.\par
%
In this paper, we propose Mask-Attention A3C (Mask A3C), which introduces an attention mechanism into Asynchronous Advantage Actor-Critic (A3C), an actor-critic-based DRL method.
Mask A3C calculates a mask-attention that is an attention map of the policy and state value, and then a visual explanation for these values is achieved by visualizing the created mask-attention.
Our method also learns the policy and state value while considering mask-attention by implementing the attention mechanism, thereby improving the performance of the agent.
The training code and evaluation code are publicly available.\footnote{\url{https://github.com/machine-perception-robotics-group/Mask-Attention_A3C}}\par

\subsubsection{Contributions}
The main contributions of this paper are as follows.
\begin{itemize}
  \item We propose a top-down visual explanation method that implements an attention mechanism in the DRL model. In the proposed method, mask-attention, which is an attention map for the outputs, can be obtained simply by forward pass.
  \item In the proposed method, the decision-making of the agent after learning can be analyzed by visualizing the acquired mask-attention. We conducted an experiment with games on the Atari 2600 and analyzed which information influences the agent's decision-making.
  \item By implementing the attention mechanism in the policy branch and value branch of the actor-critic method, a different mask-attention can be obtained depending on the policy and state value. In this way, it is possible to analyze an agent's decision-making from the two viewpoints of policy value and state value.
  \item The proposed method outputs the control value of the agent while considering mask-attention by implementing the attention mechanism. Therefore, the performance of the agent can be improved by emphasizing the area related to the control value.
\end{itemize}

\section{Related works}
\subsection{Deep reinforcement learning}
The deep Q-network (DQN) \cite{dqn-nature}, which is a typical method of DRL, expresses the action value function $Q(a|s;\theta)$ by using a neural network and acquires the optimum action by training the network parameters $\theta$.
DRL methods that learn the optimal action by a value function such as DQN are called value-based DRL, and have been studied extensively \cite{ddqn,dueling-network,c51,rainbow}.
There is also a policy-based DRL that directly learns the policy by expressing the policy $\pi(a|s;\theta)$ with a neural network \cite{ddpg,trpo,ppo,sac}.
The actor-critic method \cite{ac}, which is a policy-based method, consists of an actor that outputs the policy $\pi(a|s;\theta)$ and a critic that outputs the state value $V(s;\theta)$. 
Here, the state value $V(s;\theta)$ numerically expresses how the current state $s$ contributes to the reward.
The actor selects and performs an action according to a policy $\pi(a|s;\theta)$ that is a probability distribution from a state $s$ to an action $a$.
The critic estimates the state value $V(s;\theta)$ as the evaluation value of the policy $\pi(a|s;\theta)$ that is output by the actor.
To update the network parameters in the actor-critic method, the actor parameter update by the policy gradient method and the critic parameter update by the TD error are performed in parallel.\par
Other approaches include distributed DRL, which improves learning efficiency by constructing multiple environments and agents \cite{gorila,unreal,r2d2}.
A3C \cite{a3c} is a distributed DRL method on the basis of the actor-critic method.
A3C introduces Asynchronous, which is an asynchronous parameter update in distributed learning and Advantage, which that learns while considering rewards several steps ahead.
Experiments with the Atari 2600 showed that A3C could achieved a high score in a short training time by executing the generation of experiences used for learning in parallel.\par
In this study, we acquire mask-attention, attention maps for policy and state value, by implementing an attention mechanism in A3C.
By visualizing mask-attention at the time of inference, the decision-making of the agent acquired by learning is analyzed from the visual explanation of the policy and state value.
\subsection{Visual explanations}

\subsubsection{Visual explanations in image recognition}

In the field of image recognition, several methods utilizing an attention map have been proposed for analyzing the reason for judgments on the inference result of the network.
An attention map visualizes the network attended area at the time of inference.
Zhou \textit{et al.} proposed a class activation mapping (CAM) \cite{cam}, which acquires the attention map of a specific class from the response value of the convolutional layer and the weight of the fully connected layer.
However, the recognition performance of CAM deteriorates because it is necessary to change a part of the network structure, such as by introducing global average pooling (GAP) between the convolution and fully connected layers.
For that problem, Selvaraju \textit{et al.} proposed gradient-weighted CAM (Grad-CAM) \cite{grad-cam}, which acquires an attention map by using the response value of the convolutional layer during forward pass and the gradient during back-propagation.
Grad-CAM avoids the deterioration of the recognition performance deteriorates by generating an attention map from the gradient information. 
In image recognition, the recognition accuracy is known to improve by using an attention map during learning.
Fukui \textit{et al.} proposed an attention branch network (ABN) \cite{abn} that applies the attention map to the attention mechanism.
This method provides a visual explanation of the reason for judgment by the attention map and simultaneously improves the recognition accuracy.\par

\begin{figure*}[tb]
\centering
    \includegraphics[clip,scale=0.48]{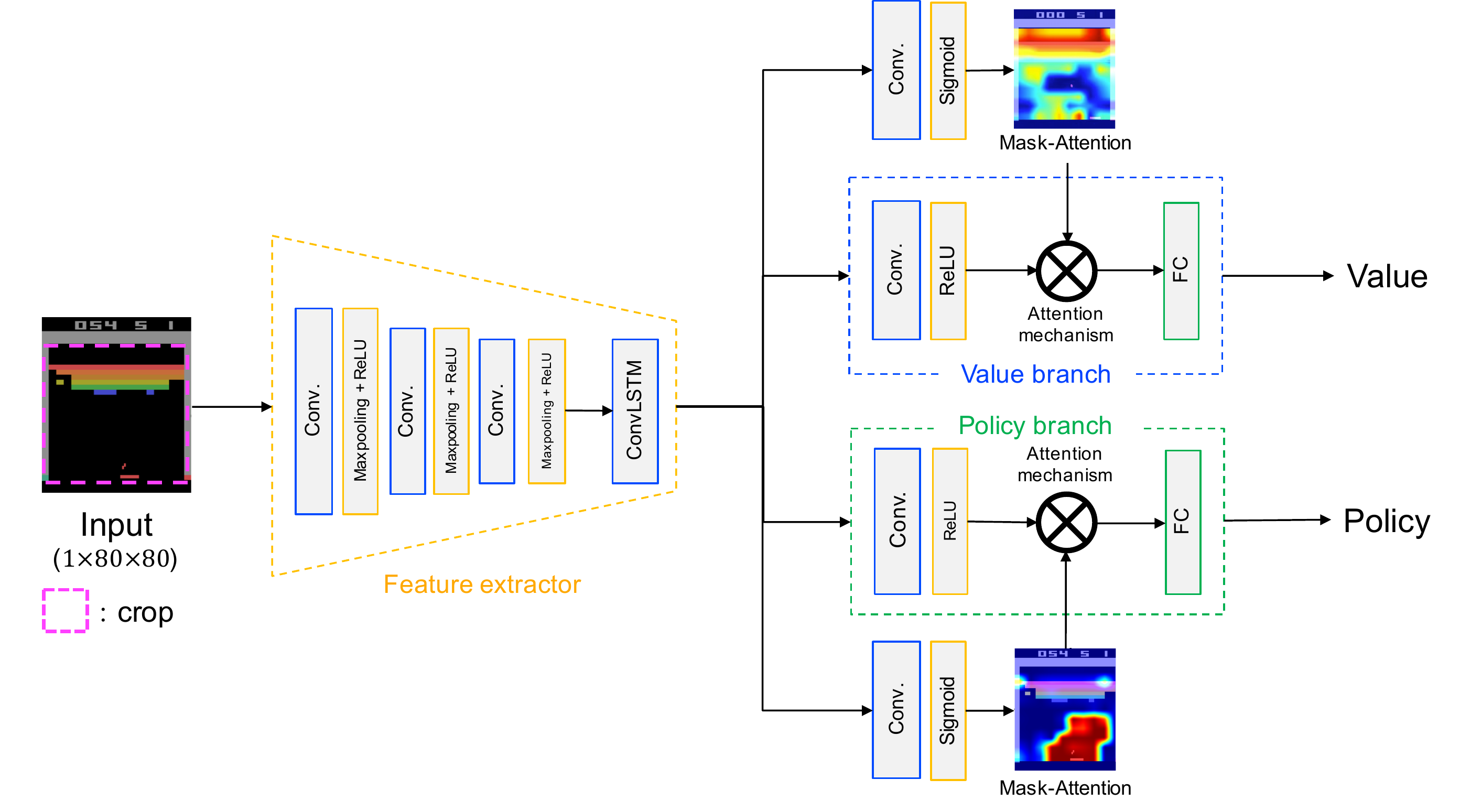}
    \caption{\textbf{Detailed network structure of Mask-Attention A3C.}} 
    \label{fig:mask-a3c}
\end{figure*}

\subsubsection{Visual explanations in deep reinforcement learning}

In DRL, several works for visual explanation of DRL models have examined.
Sorokin \textit{et al.} proposed the deep attention recurrent Q-network (DARQN) \cite{rlatt:darqn}, where an attention mechanism is implemented in DQN, a representative value-based method.
Manchin \textit{et al.} introduced a self-attention to a policy-based DRL method \cite{rlatt:manchin}, in order to improve the score along with policy analysis.
This method analyzes the agent's decision-making by using an attention map for the policy.
Our method differs in that it can improve the interpretability of the agent's decision-making in the actor-critic-based DRL method by simultaneously acquiring different attention maps for policies and state values.\par
Greydanus \textit{et al.} acquired a saliency map in A3C by calculating a perturbation image utilizing an applied Gaussian filter from the gradient during back-propagation \cite{rlatt:greydanus}.
This method takes a bottom-up approach, similar to Grad-CAM, and therefore it is necessary to perform back-propagation to acquire the saliency map.
In contrast, our method takes a top-down approach that implements an attention mechanism in the network structure of A3C.\par
Zhang \textit{et al.} proposed attention guided imitation learning (AGIL) \cite{rlatt:agil}, which guides the focus area of a network on the basis of human gaze information.
They trained a model to replicate human attention with supervised gaze heat maps. 
The input state was then augmented with this additional information. 
This style of attention fundamentally differs from that used in our work in that it incorporates hand crafted features as input.\par
The most similar work to ours was conducted by Mott \textit{et al.} \cite{rlatt:mott}.
Their method acquires two attentions (``what'' and ``where'') by using query-based attention in an actor-critic-based DRL method.
This method requires an attention query to be generated, which means significant changes must be made to the network architecture (e.g., keys, values).
In contrast, our method has a simple structure in which an attention mechanism is implemented in the policy and value branches and no significant changes to the network architecture are required.
Also, they obtained different attentions related to ``what'' and ``where'' by generating an attention query.
In contrast, our method improves interpretability by acquiring different attentions toward policy and state value, which are the outputs of actor-critic-based DRL methods.\par
The above visual explanations for DRL generates attention maps from the low-level feature maps extracted from early or middle convolutional layers, or output for action.
In DRL, a \textit{strategy} is an important clue to solve a given task and environment.
From the viewpoint of the strategy, the state value of actor-critic based model plays a crucial role because the state value is the expected value from current to future states and affects the future action selections.
However, the existing methods output attention maps with respect to the instantaneous action selection.
Our method based on an actor-critic outputs two attention maps from both the policy and value branches.
By considering both attention maps, we can clarify the basis of an agent's decision-making in more detail.
\section{Mask-Attention A3C}
We propose Mask-Attention A3C (Mask A3C), which introduces an attention mechanism into A3C, an actor-critic-based distributed DRL method.
In Mask A3C, by implementing an attention mechanism for the policy branch and the value branch, we can acquire an attention map that expresses the focus area of the network for the output of each branch.
In our study, this attention map is called mask-attention.
By visualizing the mask-attention of each branch, we obtain a visual explanation of the reason for the judgment on the policy and state value.
In addition, our method learns while considering the mask-attention by implementing the attention mechanism, which improves the performance of the agent.

\subsection{Overview of Mask A3C structure}
Figure \ref{fig:mask-a3c} shows the network structure of the proposed Mask A3C.
It consists of a feature extractor, policy branch, value branch, and attention mechanism.
Here, let $\mathbf{s}_t$ be a state at time $t$.
First, the feature extractor extracts a feature map $F_{fe}(\mathbf{s}_t)$ from the given state $\mathbf{s}_t$.
The feature extractor consists of convolutional layers.
In A3C, temporal information can be considered by utilizing LSTM, which greatly improves the performance of the agent.
However, LSTM cannot consider the spatial information of the input image, so if it is used in Mask A3C, mask-attention cannot be calculated.
We therefore utilize convolutional LSTM (ConvLSTM) \cite{convlstm}, which can consider spatiotemporal information.

%
The extracted feature map is fed into the policy and value branches.
The policy branch outputs policies, and the value branch outputs the state value function.
At these branches, the proposed method adds mask-attention module.
Each branch takes the feature map extracted from the feature extractor and computes a new feature map $F_v (\mathbf{s}_t)$ and $F_p (\mathbf{s}_t)$ by applying a convolution and a ReLU activation.
Meanwhile, the proposed method also generates mask-attention.
We denote $M_v (\mathbf{s}_t)$ and $M_p (\mathbf{s}_t)$ as the mask-attentions for value and policy branches, respectively.
The mask-attention can be generated by applying one convolutional layer of $1\times 1\times$ \# of channels and a sigmoid function to the feature map $\mathbf{s}_t$.

These feature maps and mask-attention are used for attention mechanism, the each branch output policy and state value, respectively.

\subsection{Attention mechanism}
Mask A3C implements an attention mechanism in the policy branch and value branch so that the policy and state value functions are learned in consideration of the acquired mask-attention.
The attention mechanism performs mask processing on the feature map of the middle layer in each branch by using mask-attention.
With this mask processing, the area that contributes to the optimum action and state value can be emphasized.
By using mask-attention for the feature map, the mask processing for each branch $F'_v ({\bf s}_{t})$ and $F'_p ({\bf s}_{t})$ are calculated as follows:
\begin{eqnarray}
\label{eq:mask}
F'_v ({\bf s}_{t})=F_v ({\bf s}_{t})\cdot M_v ({\bf s}_{t}), \\
F'_p ({\bf s}_{t})=F_p ({\bf s}_{t})\cdot M_p ({\bf s}_{t}).
\end{eqnarray}

The masked feature maps $F'_v ({\bf s}_{t})$ and $F'_p ({\bf s}_{t})$ are then fed into the output layer and we obtain state value and policy.
By using the masked feature map, the agent focuses on the highlighted region and selects the optimal action.

\section{Experiments}
To evaluate the effectiveness of Mask A3C, we conducted experiments using the game task of OpenAI gym \cite{openai-gym}.
Three games were used: ``Ms. Pac-Man", ``Space Invaders", and ``Seaquest".
The comparison methods were A3C, Policy Mask A3C, Value Mask A3C, and Mask A3C, for a total of four patterns.
Policy Mask A3C and Value Mask A3C refer to a Mask A3C in which the attention mechanism is implemented in only one branch (i.e., policy branch or value branch).
The learning conditions were $35$ for the number of workers, $0.0001$ for the learning coefficient, and $0.99$ for the discount rate.
The termination condition of learning was when the global steps reached $1.0\times 10^{8}$.
The termination condition of an episode was the end of one play in the game and the case where the number of steps reached $1.0\times 10^{4}$.
We used the following four evaluation methods.
\begin{itemize}
  \item Visual explanations using mask-attention
  \item Score comparison on the Atari 2600
  \item Score comparison by inverting the gaze area of mask-attention
  \item Reaction of mask-attention to new states
\end{itemize}

\subsection{Implementation details}
The input was a grayscale image of the game screen and the output was the action in each game.
The image used as input was resized to $80 \times 80$.
The output dimension of the feature extractor was 32-dimensional for the first two convolutional layers and 64-dimensional for the one remaining convolutional layer.
For the hidden state in ConvLSTM, the output dimension was 64-dimensional.
The policy branch consisted of one convolutional layer, one fully connected layer, and a softmax function.
The output dimension of the convolution layer was 32-imensional and the number of output units of the fully connected layer was the number of actions in each game.
The value branch consisted of one convolutional layer with a 32-dimensional output dimension and one fully connected layer with one output unit.
The network structure of A3C in this experiment was the structure excluded the attention mechanism from Mask A3C.

\begin{figure*}[t!]
\begin{tabular}{cc}
\begin{minipage}[t]{1.0\hsize}
\centering
    \includegraphics[clip,scale=0.4]{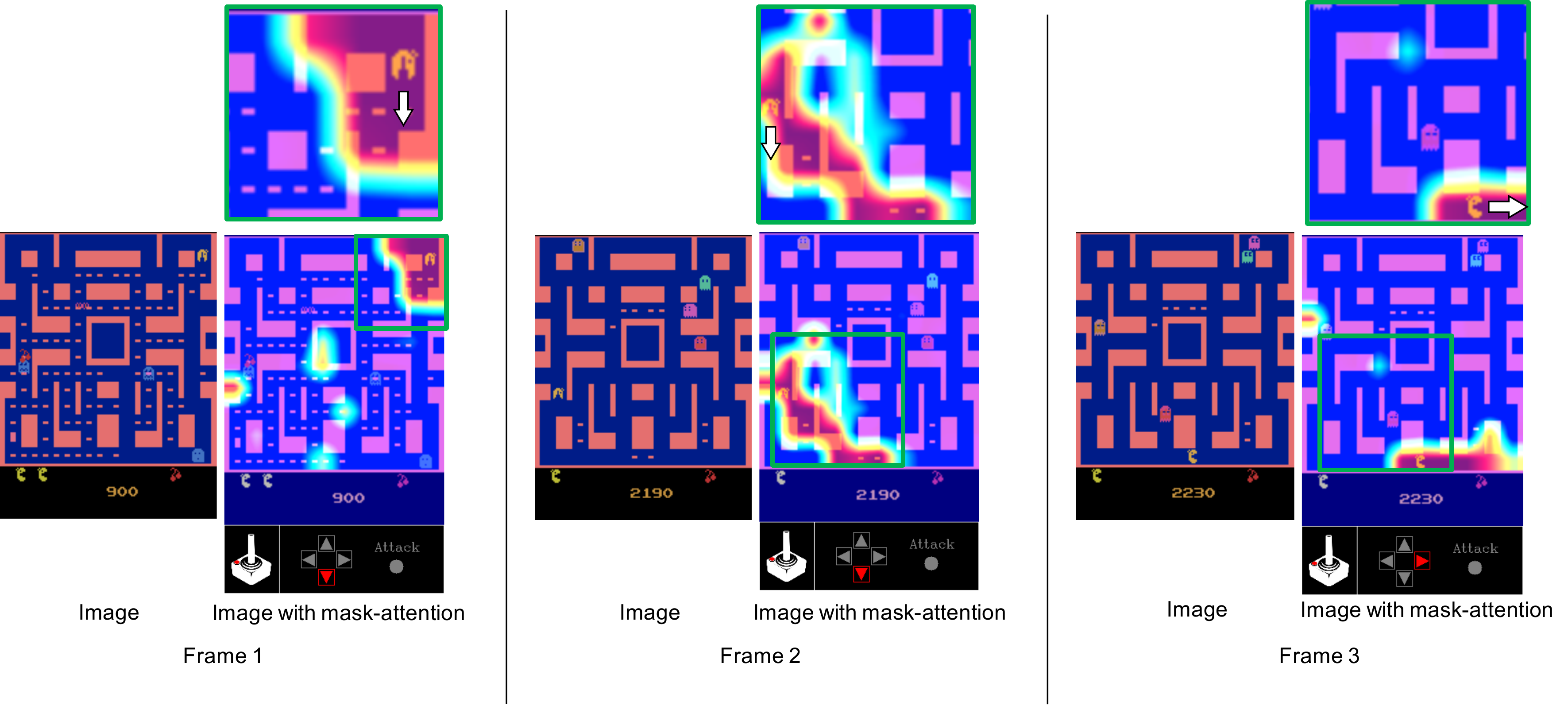}
\subcaption{\textbf{Ms. Pac-Man}: The white arrow shows the direction of travel of Pac-Man.}
\vspace{0.2cm}
\label{fig:visual-mspacman-policy}
\end{minipage} \\
\begin{minipage}[t]{1.0\hsize}
\centering
    \includegraphics[clip,scale=0.4]{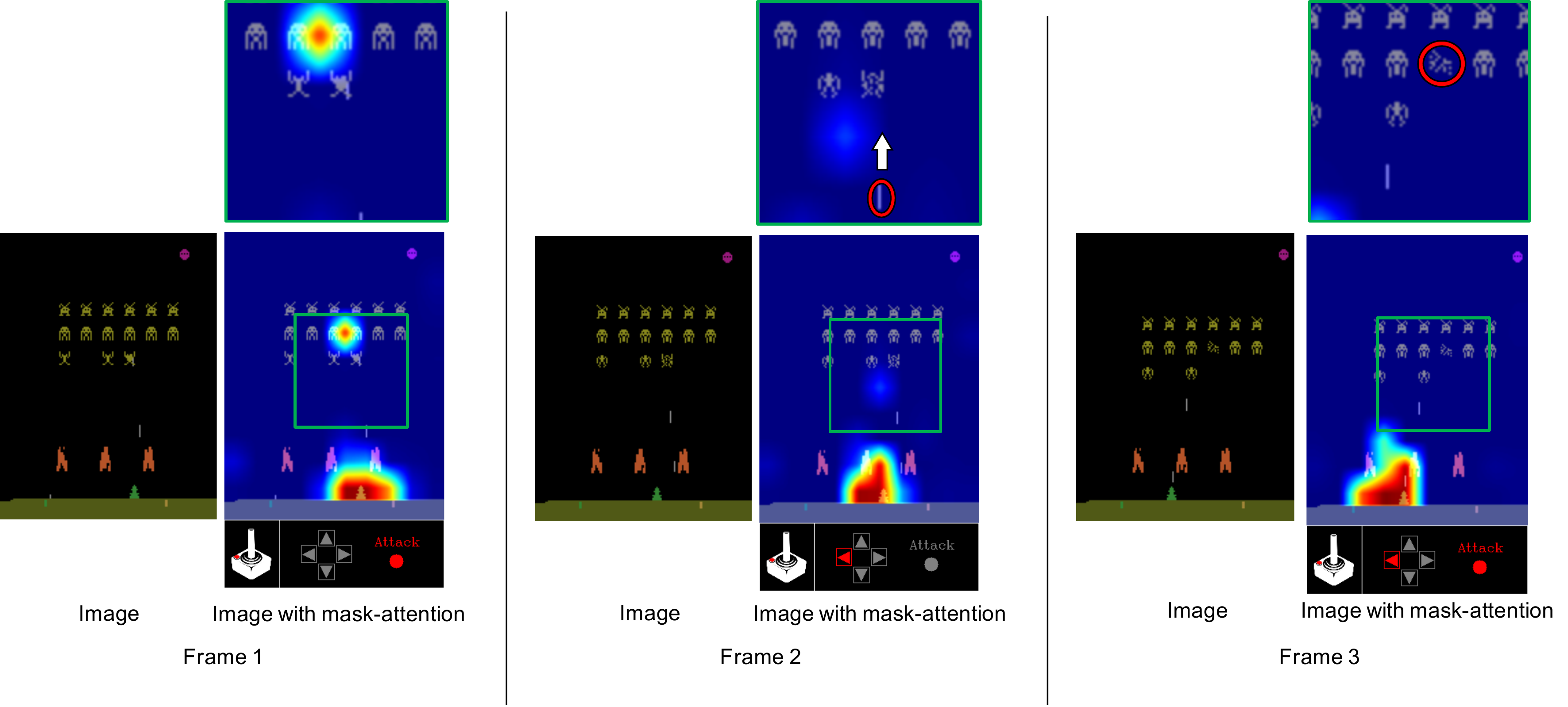}
\subcaption{\textbf{Space Invaders}: The red circles in Frame 2 show the beam that is the attack of the agent in Frame 1, and the red circles in Frame 3 show the destroyed invaders. The white arrow shows the direction of travel of the beam that is the attack of the agent.}
\vspace{0.2cm}
\label{fig:visual-spaceinvaders-policy}
\end{minipage}\\
\begin{minipage}[t]{1.0\hsize}
\centering
        \includegraphics[clip,scale=0.4]{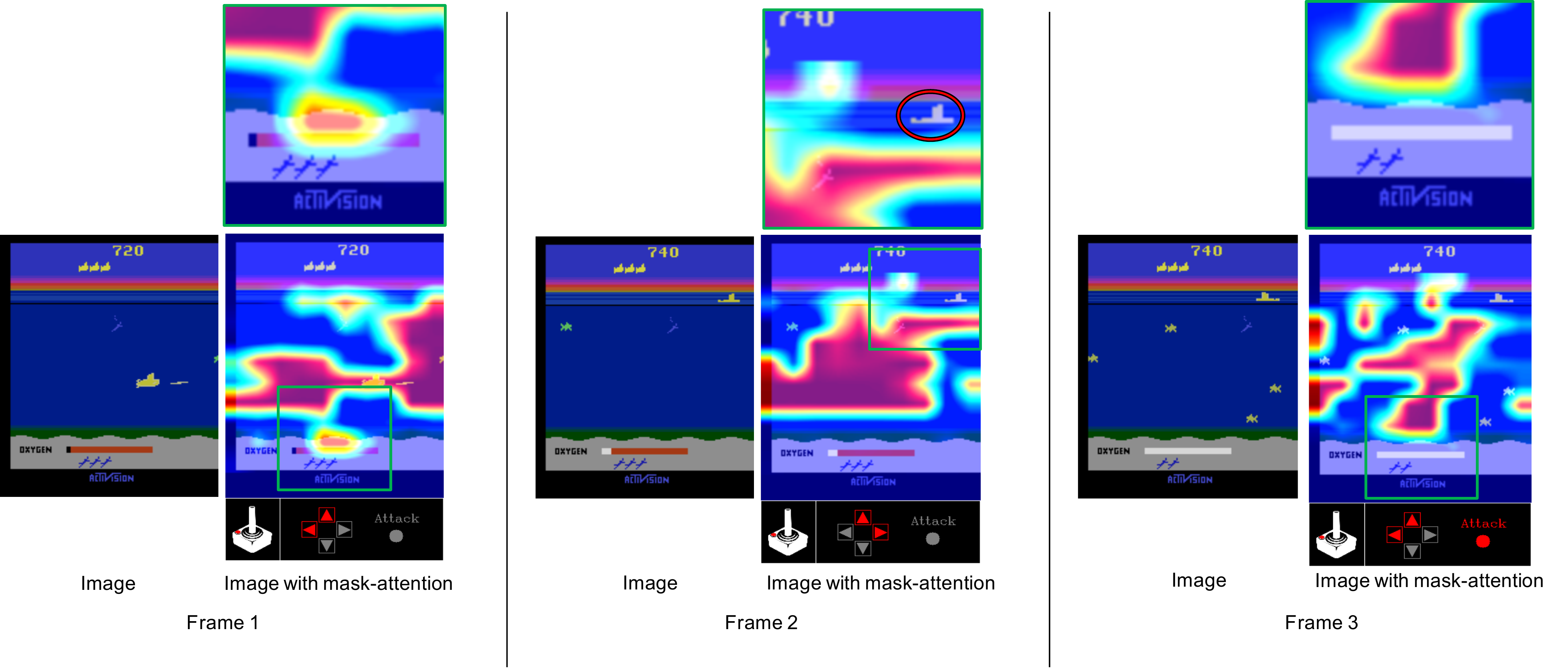}
\subcaption{\textbf{Seaquest}: The red circle in Frame 2 shows the submarine replenishing oxygen.}
\label{fig:visual-seaquest-policy}
\end{minipage}
\end{tabular}
\caption{\textbf{Visualization example of mask-attention in policy}: The controller in ``Image with mask-attention" is an action that is output by the DRL model. The green broken line in the State value shows the transition to the next stage.}
\label{fig:visual-policy}
\end{figure*}

\begin{figure*}[t!]
\begin{tabular}{cc}
\hspace{-0.3cm}
\begin{minipage}[t]{1.0\hsize}
\centering
    \includegraphics[clip,width=\linewidth]{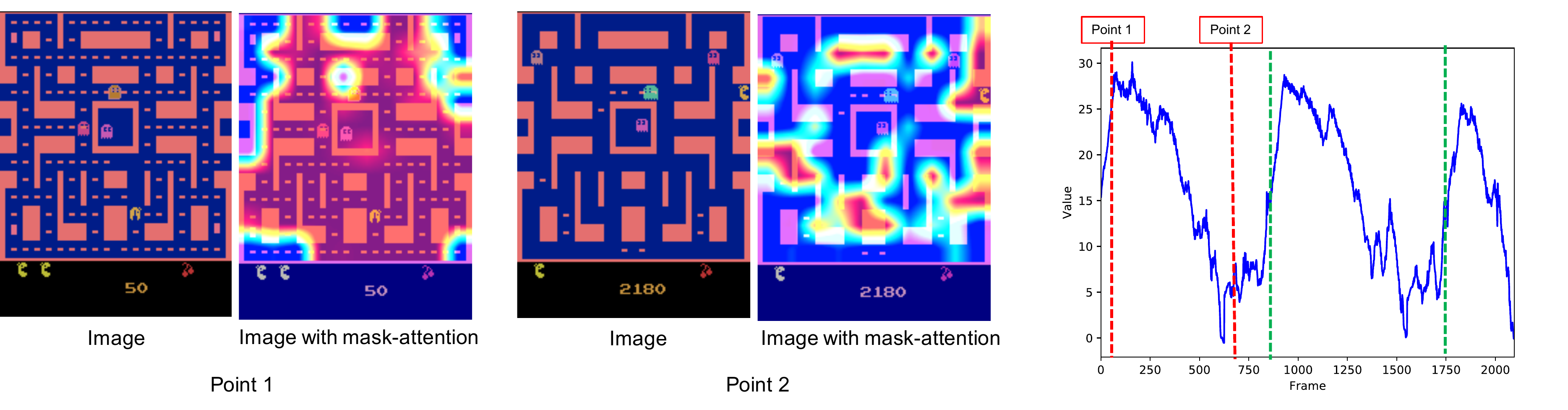}
\vspace{-0.6cm}\subcaption{\textbf{Ms. Pac-Man}}
\label{fig:visual-mspacman-value}
\end{minipage} \\
\hspace{-0.3cm}
\begin{minipage}[t]{1.0\hsize}
\centering
    \includegraphics[clip,width=\linewidth]{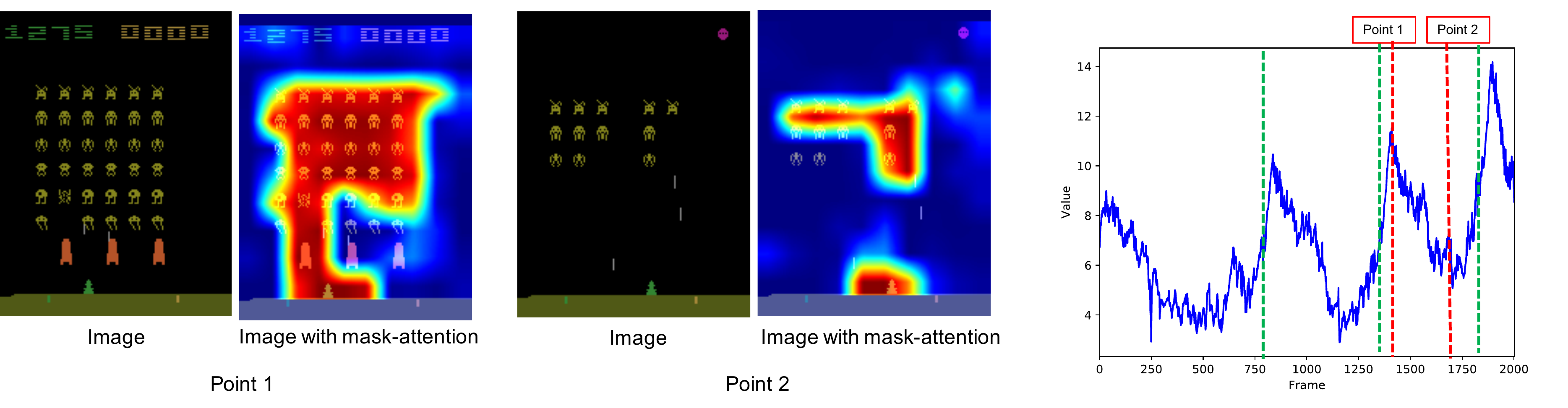}
\vspace{-0.6cm}\subcaption{\textbf{Space Invaders}}
\label{fig:visual-spaceinvaders-value}
\end{minipage}\\
\hspace{-0.3cm}
\begin{minipage}[t]{1.0\hsize}
\centering
    \includegraphics[clip,width=\linewidth]{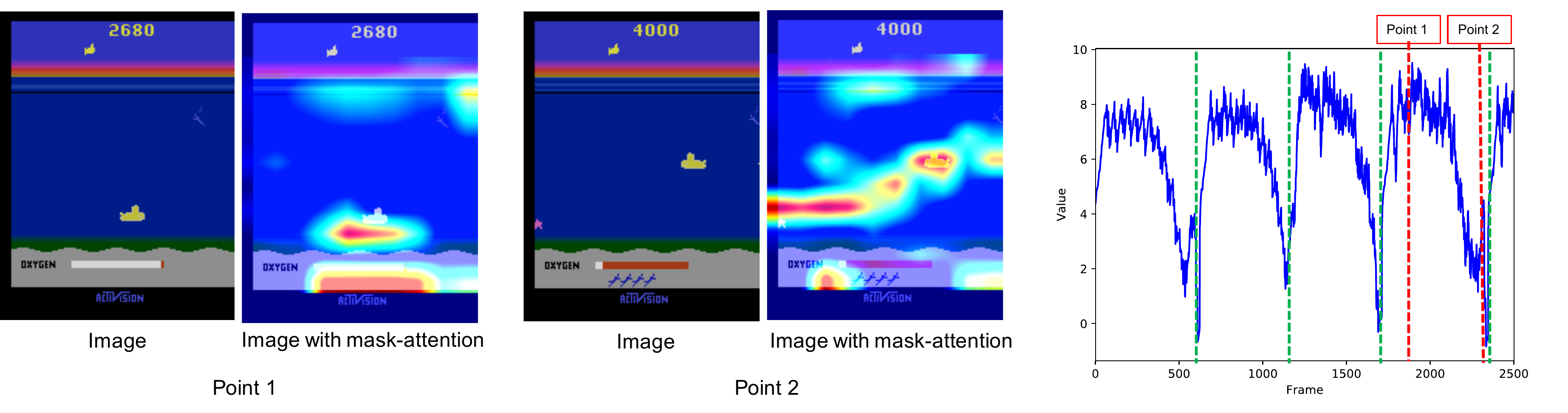}
\vspace{-0.6cm}
\subcaption{\textbf{Seaquest}}
\label{fig:visual-seaquest-value}
\end{minipage}
\end{tabular}
\caption{\textbf{Visualization example of mask-attention in state value}: The green broken line in the graph shows the transition to the next stage.}
\label{fig:visual-value}
\end{figure*}

\subsection{Visual explanations using mask-attention}
Figures \ref{fig:visual-policy} and \ref{fig:visual-value} show visualization examples of mask-attention in Atari 2600. (Examples of other frames and other environment are available in Appendix \ref{appendix:a}.)
Hereafter, we discuss the obtained mask-attentions shown in these figures.\par

\subsubsection{Ms. Pac-Man}
Ms. Pac-Man is a game in which the player collects scattered cookies while avoiding enemies.
The actions of the agent that is the Pac-Man are ``Noop", ``Up", ``Down", ``Left", ``Right", ``Up + Left", ``Up + Right", ``Down + Left", and ``Down + Right".
In Fig. \ref{fig:visual-policy}(a), the agent of Frame 1 was attending around Pac-Man.
The agent of Frame 2 was attending to the cookies remaining on the screen.
In Frame 3, Pac-Man moved to the gazed area of Frame 2 and acquired the cookies.
These results demonstrate that the agent controlled Pac-Man toward the cookies while simultaneously attending to the surroundings of Pac-Man.
%
In Fig. \ref{fig:visual-value}(a), Point 1, which was the beginning of the game, the agent was attending to the entire screen.
In contrast, Point 2 is reduced the gaze area in accordance with the decreases of cookies.
Also, from Point 1 to Point 2, the state value decreased as the cookies on the screen decreased.
These results demonstrate that the agent recognized the cookies as the score source.\par
%
%


\subsubsection{Space Invaders}
Space Invaders is a shooting game in which the player repels the enemy invaders.
The actions of the agent that is the cannon are ``Noop", ``Left", ``Right", ``Attack", ``Left + Attack", and ``Right + Attack".
In Fig. \ref{fig:visual-policy}(b), the agent in Frame 1 is attended to the invaders and the action was ``Attack".
In Frame 2, we can see that the beam in Frame 1 was heading toward the invader, and in Frame 3, the beam was repelling the invaders that were attended to in Frame 1.
In all frames, the agent attended around itself while avoiding the defensive walls.
From these results, we can see that the agent repelled the invaders while simultaneously avoiding the defensive wall.
In Fig. \ref{fig:visual-value}(b), the agent in Point 1 was attending to all of the invaders and the agent in Point 2 was shrinking the gaze area in accordance with the decreasing number of invaders.
In addition, from Point 1 to Point 2, the state value is decreased according to the number of invaders.
These results demonstrate that the agent recognized the invaders as the score source.\par
%
%



\subsubsection{Seaquest}
Seaquest is a game in which players use submarines to rescue divers and destroy enemies and fish. 
The actions of the agent that is the submarine are ``Noop", ``Up", ``Down", ``Left", ``Right", and ``Attack".
%
In Fig. \ref{fig:visual-policy}(c), the agent in Frame 1 is attending to the oxygen gauge and the action was ``Up".
In Frame 2, the submarine operated by the agent is floating on the surface of the sea.
When the amount of oxygen is reduced in Seaquest, it can be replenished when the submarine rises to the surface of the sea.
In Frame 3, we can see that the oxygen gauge is full due to the rise to the sea level in Frame 2, and there is no gaze on the oxygen gauge.
From these results, we can see that the agent recognizes the oxygen gauge is low and controls the submarine to rise to sea level.
In Fig. \ref{fig:visual-value}(c), the agent in Point 1 was attending to the entire oxygen gauge.
On the other hand, in Point 2, where the amount of oxygen is low, the gaze area of the oxygen gauge was reduced in accordance with the amount of oxygen.
The state value also decreased from Point 1 to Point 2. 
These results demonstrate that the agent recognized the oxygen as the score source.\par
%
%


\begin{table*}[t]
    \begin{center}
		\caption{\textbf{Max and mean scores over 100 episodes on Atari 2600}: Scores of models that had the highest average score among five trials in each method are shown.}
 		\scalebox{1.15}{
		\begin{tabular}{cc||cc|cc|cc|cc|cc|cc} \hline
			     \multicolumn{2}{c||}{Att. mechanism} & \multicolumn{2}{c|}{Breakout} & \multicolumn{2}{c|}{Ms. Pac-Man} & \multicolumn{2}{c|}{Space Invaders} &
			     \multicolumn{2}{c|}{Beamrider} &
			     \multicolumn{2}{c|}{Fishing Derby} &
			     \multicolumn{2}{c}{Seaquest} \\  \cline{3-14}
			    Policy & Value & max & mean & max & mean & max & mean & max & mean & max & mean & max & mean \\ \hline \hline
			    &  & {\bf 864} & {\bf 662.0} & 5380 & 4573.3 & 19505 & 18531.8 & 34748 & 28341.1 & 41 & 32.1 & 2760 & 2728.2 \\ \hline
			    \checkmark  &  & {\bf 864} & 595.8 & 6330 & 4833.8 & {\bf  19860} & 19102.8 & 32604 & {\bf 28495.3} & 41 & {\bf 37.5} & 2820 & 2784.0 \\ \hline
			    & \checkmark & {\bf 864} & 606.9 & 4830 & 4044.5 & 19675 & 18537.8 & {\bf 35108} & 28205.7 & {\bf 43} & 36.1 & 2820 & 2786.4 \\ \hline
			    \checkmark  & \checkmark & {\bf 864} & 640.0 & {\bf 6610} & {\bf 5314.1} & 19810 & {\bf 19212.5} & 34448 & 27671.1 & 41 & 34.3 & {\bf 17150} & {\bf 6701.9} \\ \hline
                \end{tabular}
                }
                \label{table:score}
	\end{center}
\end{table*}

\begin{table*}[t]
    \begin{center}
		\caption{\textbf{Score comparison by inverting the gaze area of mask-attention}: Normal and inverse are the scores when the gaze area was not inverted and is inverted, respectively. Random is the score when the action was randomly selected. Max / mean = maximum and average scores over 100 episodes.}
 		\scalebox{1.06}{
		\begin{tabular}{cc||c||cc|cc|cc|cc|cc|cc} \hline
			     \multicolumn{2}{c||}{Att. mechanism} &  & \multicolumn{2}{c|}{Breakout} & \multicolumn{2}{c|}{Ms. Pac-Man} & \multicolumn{2}{c|}{Space Invaders} &
			     \multicolumn{2}{c|}{Beamrider} &
			     \multicolumn{2}{c|}{Fishing Derby} &
			     \multicolumn{2}{c}{Seaquest} \\ \cline{4-15} 
			    Policy & Value &  & max & mean & max & mean & max & mean & max & mean & max & mean & max & mean \\ \hline \hline
			    \multirow{2}{*}{\checkmark} & & normal & 864 & 595.8 & 6630 & 4833.8 & 19860 & 19102.8 & 32604 & 28495.3 & 41 & 37.5 & 2820 & 2784.0 \\ \cline{3-15}
			    & & inverse & 4 & 2.2 & 290 & 268.9 & 805 & 306.9 & 4996 & 1554.2 & -49 & -75.7 & 280 & 158.2 \\ \hline
			    \multirow{2}{*}{\checkmark} & \multirow{2}{*}{\checkmark} & normal & 864 & 640.0 & 6610 & 5314.1 & 19810 & 19212.5 & 34448 & 27671.1 & 41 & 34.3 & 17150 & 6701.9 \\ \cline{3-15}
			    & & inverse & 5 & 1.8 & 410 & 194.4 & 915 & 420.2 & 6380 & 2063.9 & -49 & -74.7 & 420 & 280.6 \\ \hline
			    \multicolumn{3}{c||}{random} & 5 & 1.2 & 1080 & 247.8 & 460 & 142.1 & 852 & 356.5 & -85 & -93.1 & 300 & 82.8 \\ \hline
    \end{tabular}
    }
    \label{table:inverse}
	\end{center}
    \begin{center}
		\caption{\textbf{Decrease rate of score due to inverse of gaze area in mask-attention ($\%$)}: Max / mean = maximum and average scores over 100 episodes.}
 		\scalebox{1.2}{
		\begin{tabular}{cc||cc|cc|cc|cc|cc|cc} \hline
			     \multicolumn{2}{c||}{Att. mechanism} & \multicolumn{2}{c|}{Breakout} & 
			     \multicolumn{2}{c|}{Ms. Pac-Man} & 
			     \multicolumn{2}{c|}{Space Invaders} &
			     \multicolumn{2}{c|}{Beamrider} &
			     \multicolumn{2}{c|}{Fishing Derby} &
			     \multicolumn{2}{c}{Seaquest} \\ \cline{3-14} 
			    Policy & Value & max & mean & max & mean & max & mean & max & mean & max & mean & max & mean \\ \hline \hline
			    \checkmark &  & 99.53 & 99.63 & 95.41 & 94.43 & 95.94 & 98.39 & 84.67 & 94.54 & 98.90 & 99.12 & 90.07 & 94.31 \\ \cline{3-14} \hline
			    \checkmark & \checkmark & 99.42 & 99.71 & 93.79 & 96.34 & 95.38 & 97.80 & 81.47 & 92.54 & 98.90 & 99.09 & 97.55 & 95.81 \\ \cline{3-14} \hline
                \end{tabular}
                }
                \label{table:inverse-per}
	\end{center}
\end{table*}

\subsubsection{Discussion}
We obtained different mask-attentions in the policy and state value by implementing an attention mechanism for each branch.
The policy represents the probability distribution of the action selection in the current state.
Therefore, the mask-attention of the policy indicates the area that contributes to the action of the agent.
The state value represents the expected value of the return in the current state.
Here, return is the sum of rewards during one episode.
Therefore, the mask-attention of the state value indicates the area that represents the property of the game.

\subsection{Score comparison on the Atari 2600}
Table \ref{table:score} shows the max and mean scores over 100 episodes on the Atari 2600.
As we can see, the mean score in Breakout was lower for Policy Mask A3C, Value Mask A3C, and Mask A3C than for A3C.
In contrast, the max score in Breakout was 864 for all methods.
This score is the best score that can be obtained in Breakout.
Breakout is a simple game  with no external factors: it simply consists of the player hitting the ball back with a paddle.
Therefore, we presume that A3C and Mask A3C got the same score.
In Ms. Pac-Man, the max and mean scores of Policy Mask A3C and Mask A3C improved compared to those of A3C.
The control of the agent in Ms. Pac-Man is complex because it is necessary to select an action while considering external factors (e.g., the enemy).
Policy Mask A3C and Mask A3C, which implement an attention mechanism on the policy branch, can emphasize the areas that contribute to the action (e.g., cookies and enemies), which is why Policy Mask A3C and Mask A3C obtained a higher score than A3C.
In Space Invaders, Policy Mask A3C and Mask A3C improved the max and mean scores compared to A3C.
Also, the max and mean scores of Value Mask A3C were almost the same as those of A3C.
The agent in Space Invaders needs to select actions in consideration of external factors (e.g., enemies), the same as in Ms. Pac-Man.
Policy Mask A3C and Mask A3C, which implement an attention mechanism in the policy branch, can emphasize the areas that contribute to the action (e.g., defensive walls and invaders), which is why they obtained a higher score than A3C.
There was no significant difference among the scores for Beamrider.
In Beamrider, two kinds of enemies exist from the agent's point of view: one, an enemy that the agent should defeat, and two, the necessity that the agent should avoid collisions. 
These enemies are like similar in appearance.
Since our attention mechanism and the mask-attention that highlights the enemies are insufficient to distinguish these enemies, they do not contribute to the score improvement.
In Fishing Derby, Policy Mask A3C, Value Mask A3C, and Mask A3C improved the mean scores compared to A3C.
In Fishing Derby, there are many fish that are score sources, and getting the closest fish is the fastest way to get a point.
Policy Mask A3C, Value Mask A3C, and Mask A3C, which implements an attention mechanism, can emphasize the fish closest to the player.
This is why the mask-attention methods obtained a higher score than that A3C.
In Seaquest, Mask A3C improved both the maximum and average scores compared to the other methods. 
This is because only the Mask A3C agent could learn the action of replenishing oxygen. 
Seaquest features an oxygen gauge at the bottom of the screen, and the game ends when the oxygen is gone. 
Mask A3C, which implements an attention mechanism in the policy and value branches, can emphasize the oxygen gauge, which is why Mask A3C obtained a higher score than the other methods.

\subsection{Comparison of scores by inverting gaze area in mask-attention}
In visually explaining the decision-making of an agent using Mask A3C, we want to verify whether mask-attention represents an effective gaze area for the optimum action.
In a case where the game score obtained by inverting mask-attention does not change, it means the mask-attention does not contribute to the agent's action.
In contrast, if the game score significantly decreases, it means the mask-attention largely contributes to the agent's actions acquiring the game score.
For this verification method, we created a map in which the gaze area of the mask-attention in the policy branch is inverted and then calculated the score on Atari 2600 when the created map was used for the attention mechanism.
We investigated whether mask-attention is effective for the visual explanation of an action by comparing the scores when the gaze area was inverted and when it was not inverted.
The map in which the gaze area of the mask-attention is inverted was created by
\begin{eqnarray}
  \label{eq:inverse}
  M_{\rm inverse}({\bf s}_{t})=1-M({\bf s}_{t}) ,
\end{eqnarray}
where ${\bf s}_{t}$ is the state (grayscale image in the experiment), $M({\bf s}_{t})$ is mask-attention, and $M_{\rm inverse}({\bf s}_{t})$ is the map after inverting the gaze area of mask-attention.\par
Table \ref{table:inverse} shows the score comparison by inverting the gaze area of mask-attention. 
(Also, Table \ref{table:inverse-per} shows the decrease ratio from the normal score to the inverse score.)
As shown in Table \ref{table:inverse}, the inverse score was significantly lower than the normal score in all games.
In addition, the inverse score in Breakout was equivalent to a random score, and the inverse score of Mask A3C in Ms. Pac-Man was 53.4 lower than random.
In contrast, we can see that the inverse score in Space Invaders, Beamrider, Fishing Derby, and Seaquest was higher than a random score.
However, from Table \ref{table:inverse-per}, we can see that the decrease ratio in the mean of Space Invaders, Beamrider, Fishing Derby, and Seaquest was more than 90\%, as with the other games.
Therefore, we conclude that the gaze area of mask-attention in the policy branch can represents a useful area for action to obtain a high score.
%
\subsection{Reaction of mask-attention to new states}
We want to verify whether mask-attention and the behavior of an agent are affected when changes are made to the object that mask-attention was attending to.
In a case where the behavior of the agent significantly changes  by making changes to the object being attended, the object is an important component that contributes to the agent's behavior.
In this experiment, we focused on the fish and oxygen gauges in Seaquest.
For this verification method, we added the fish to the image when evaluating Mask A3C in Seaquest.
The frame to add fish is the frame in which the agent destroys the fish and the fish disappears from the image.
After adding the fish to the image, we investigated the behavior of the agent and the change in mask-attention.
Similarly, we added the oxygen gauge full of oxygen to the image.
The frame to add the oxygen gauge is before the agent replenishes oxygen.\par
\begin{figure*}[p]
\begin{tabular}{cc}
\begin{minipage}[b]{1.0\hsize}
\centering
    \includegraphics[clip,width=\linewidth]{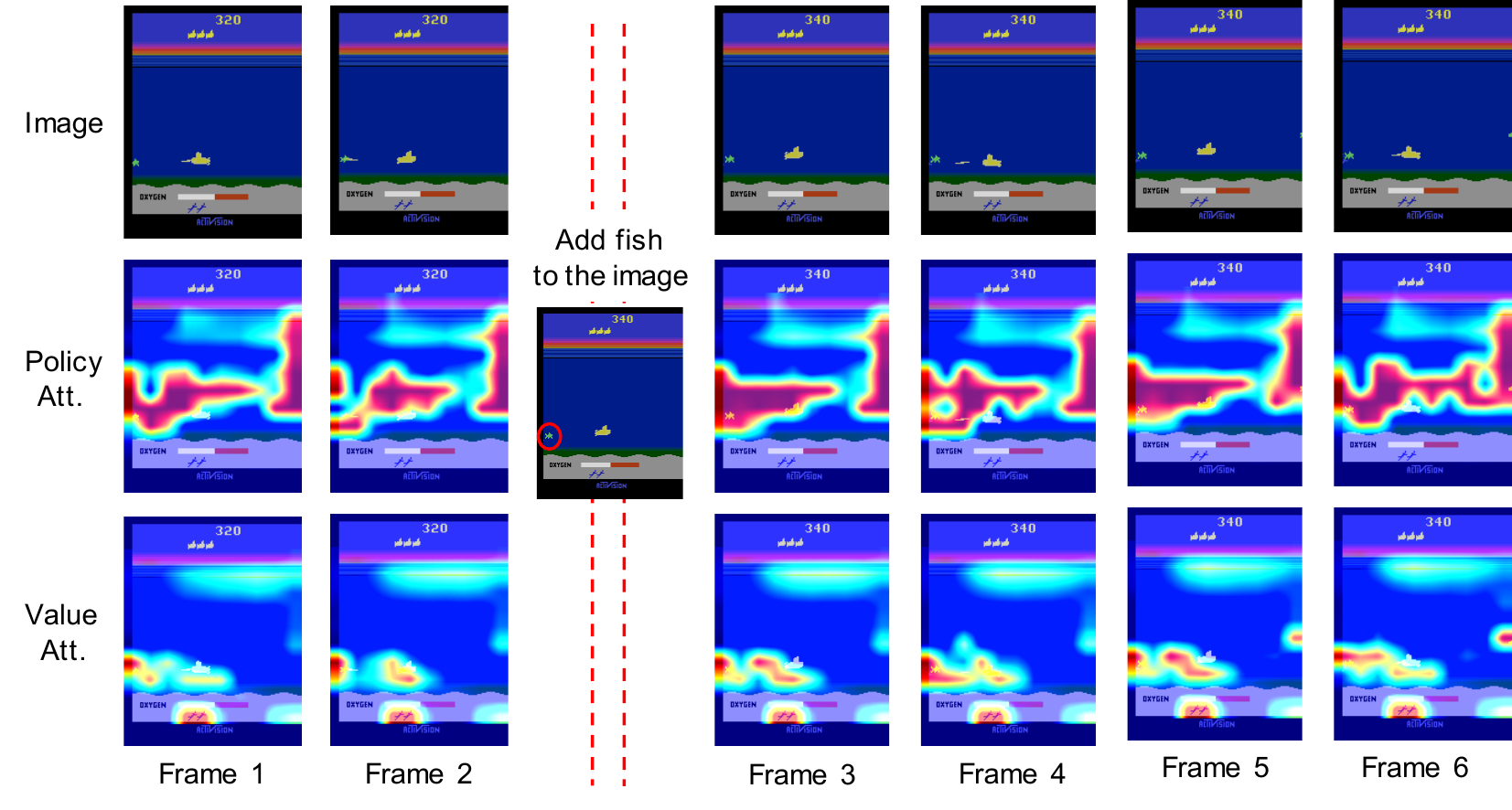}
\subcaption{\textbf{Reaction to fish-related new states}}
\vspace{0.2cm}
\label{fig:add-fish}
\end{minipage} \\
\begin{minipage}[t]{1.0\hsize}
\centering
    \includegraphics[clip,width=\linewidth]{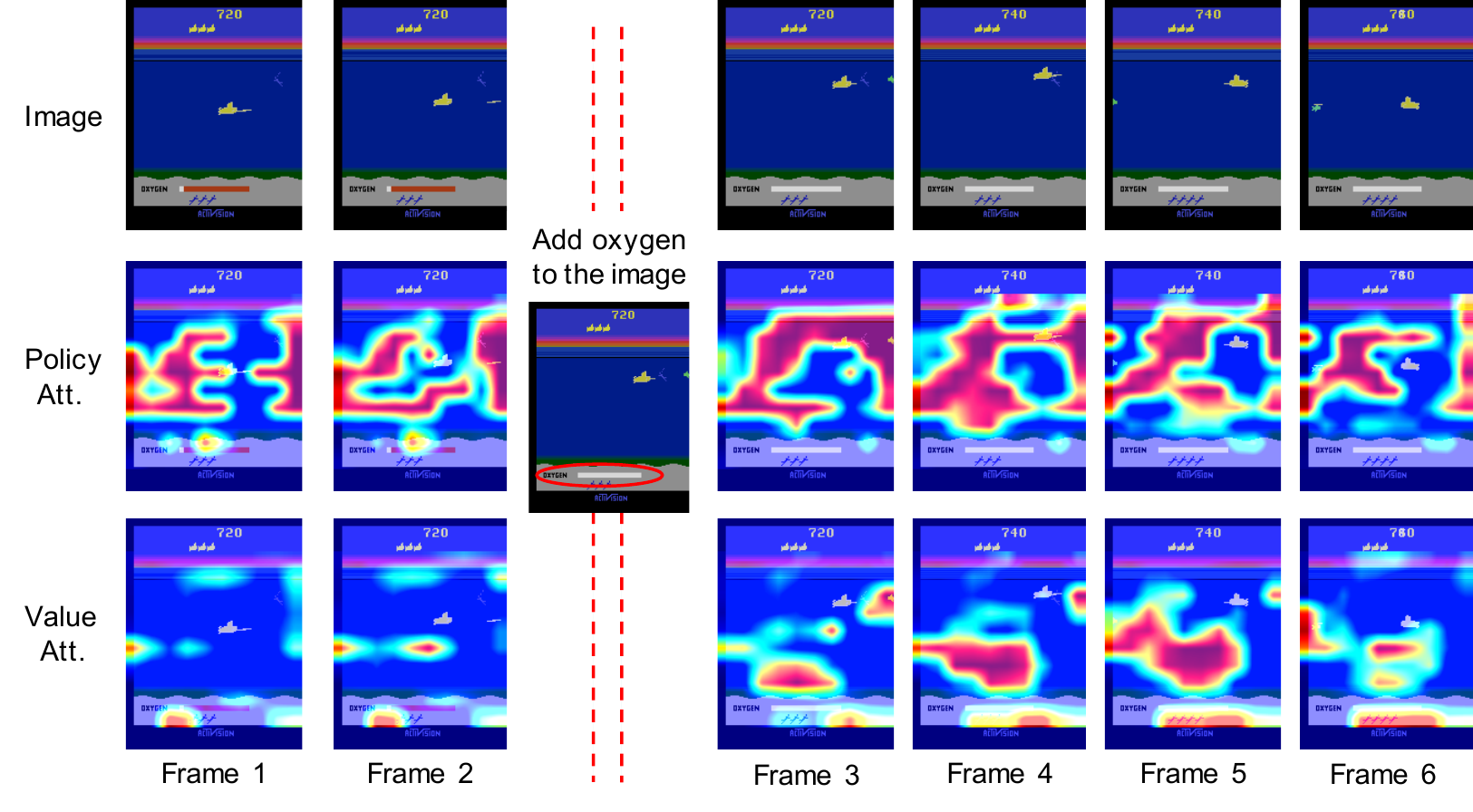}
\subcaption{\textbf{Reaction to oxygen-related new states}}
\label{fig:add-oxygen}
\end{minipage}
\end{tabular}
\caption{\textbf{Visualization example of mask-attention to new states}}
\label{fig:add-object}
\end{figure*}

%
Figure \ref{fig:add-fish} shows the changes in agent behavior and mask-attention due to the addition of fish.
Here, the fish was added after Frame 2. 
From the Figure \ref{fig:add-fish}, we can see that the policy / value mask-attention after Frame 3 is attending to the fish.
At this time, the agent is moving to the right in Frame 3, but is attacking the fish added in Frame 4.
These results demonstrate that the fish in Seaquest is an object that greatly contributes to the behavior of the agent.  
Figure \ref{fig:add-oxygen} shows changes in agent behavior and mask-attention due to the addition of oxygen gauge.
Here, the oxygen gauge is added after Frame 2. 
From the figure \ref{fig:add-oxygen}, we can see that the policy mask-attention after Frame 3 is not attending to the oxygen gauge.
Also, the value mask-attention of Frames 3 and 4 in Fig. \ref{fig:add-oxygen} is not attending to the entire oxygen gauge, unlike Fig. \ref{fig:visual-seaquest-value} Point 1.
However, the value mask-attention of Frames 5 and 6 is attending to the entire oxygen gauge, similar to Fig. \ref{fig:visual-seaquest-value} Point 1.
At this time,, the agent's behavior rose to sea level in Frames 3 and 4, but descended in Frames 5 and 6.
These results demonstrate that the amount of oxygen in the oxygen gauge in Seaquest is an object that greatly contributes to the behavior of the agent.\par
%
The policy mask-attention in the experiment of adding the oxygen gauge is not attending to the oxygen gauge in Frame 3.
Here, Frame 3 is the frame immediately after adding the oxygen gauge.
In other words, the policy mask-attention immediately reflects the effect of adding the oxygen gauge. 
On the other hand, in the value mask-attention in the same experiment, the entire oxygen gauge is not attended to in Frames 3 and 4, but it is attended to in Frames 5 and 6. 
These results demonstrate that the change in mask-attention due to the addition of an oxygen gauge is different for the policy and value mask-attention. 
Therefore, we conclude that policy mask-attention shows the area that contributes to the behavior in the current state, and value mask-attention shows the area related to the characteristic of the game considering the time series.

\section{Conclusion}
In this paper, we proposed Mask-Attention A3C (Mask A3C), which introduces an attention mechanism into Asynchronous Advantage Actor-Critic (A3C).
In Mask A3C, we acquire a mask-attention that expresses the important area for the policy and state value by implementing an attention mechanism in the policy and value branches.
This enables a visual explanation of the judgment reason in the decision-making of the agent, from the two viewpoints of policy and state value, by visualizing mask-attention.
We also emphasize which areas contribute to the optimal action and state value by implementing an attention mechanism, which simultaneously improves the performance of the agent.\par
Experiments with the Atari 2600 confirmed the acquisition of different mask-attentions in the policy and value branches.
The results demonstrate that the mask-attention of the policy branch indicates the area that contributes to the action while that of the value branch indicates the area that expresses the property of the game.
We provided a useful analysis for the decision-making of agents in game tasks from the two viewpoints of policy and state value by visualizing these mask-attentions.
A comparison of game scores showed that the score improved when the attention mechanism was implemented in the policy branch.
However, our experiments were conducted with game tasks that are easy to analyze visually.
Our future work will entail the visual analysis of agents using mask-attention is a future work for complex tasks (e.g., robot control and autonomous driving).

\section*{Acknowledgment}
This paper is based on results obtained from a project, JPNP20006, commissioned by the New Energy and Industrial Technology Development Organization (NEDO).

\bibliographystyle{IEEEtran}
\bibliography{arxiv}

\newpage

\onecolumn

\section*{{\LARGE Appendix}}
\label{appendix:a}

\subsection{Additional mask-attention examples}
\subsubsection{\textbf{Mask-attentions in policy}}
\begin{figure*}[h!]
	\centering
    \includegraphics[clip,width=\linewidth]{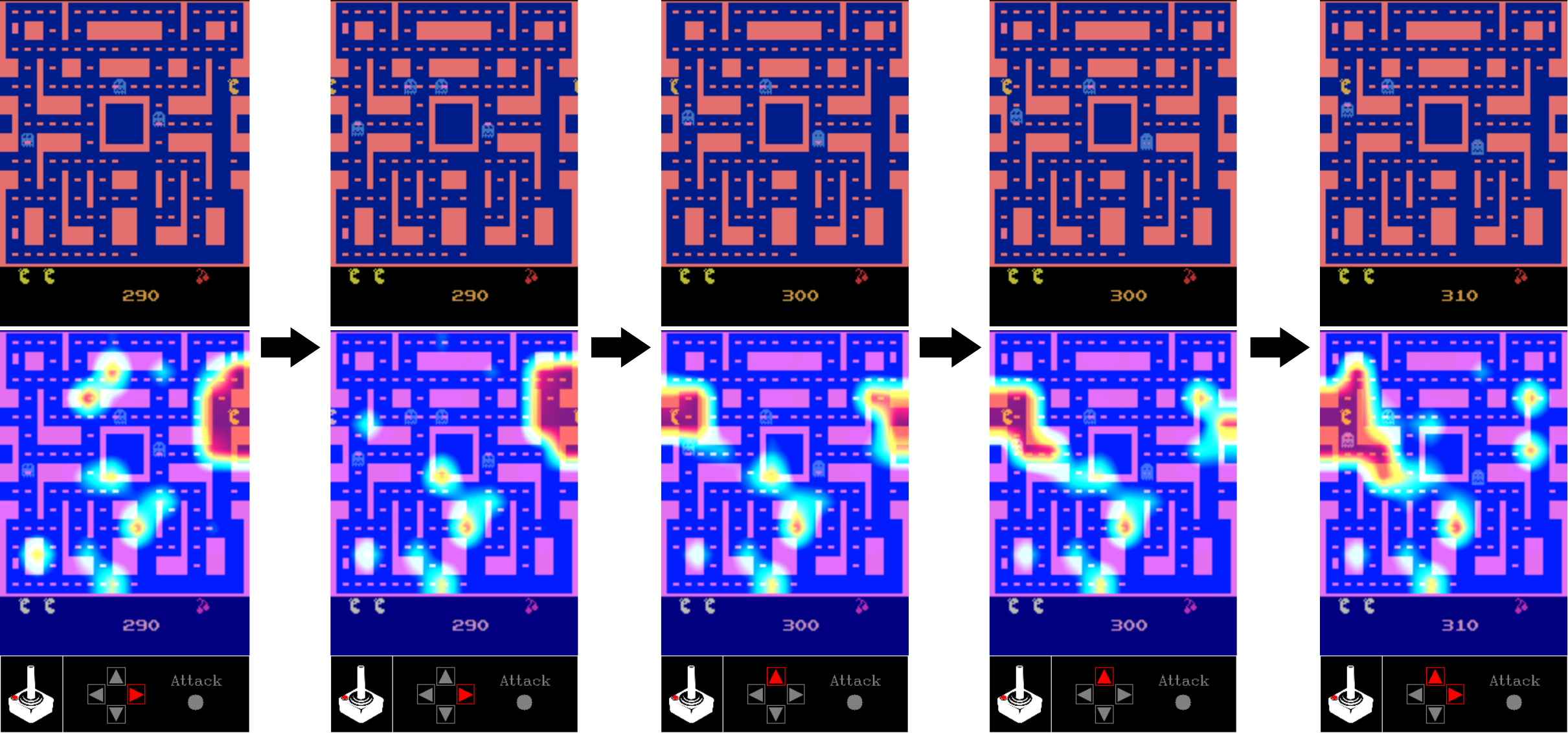}
	\caption{\textbf{Mask-attention in policy of Ms.Pac-Man.}}
	\label{fig:appendix-policy-ms}
\end{figure*}

\begin{figure*}[h!]
	\centering
    \includegraphics[clip,width=\linewidth]{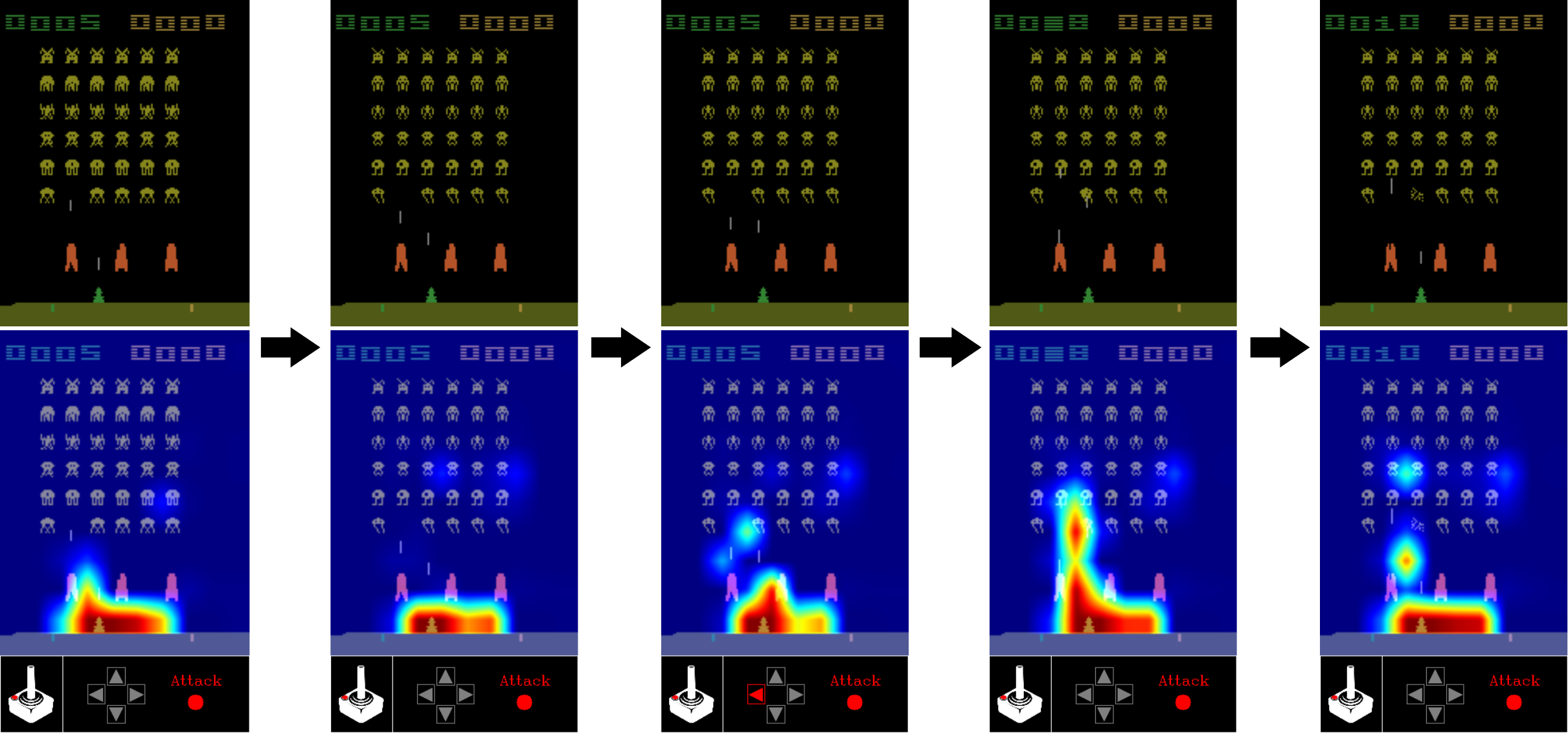}
	\caption{\textbf{Mask-attention in policy of Space Invaders.}}
	\label{fig:appendix-policy-si}
\end{figure*}

\begin{figure*}[h!]
	\centering
    \includegraphics[clip,width=\linewidth]{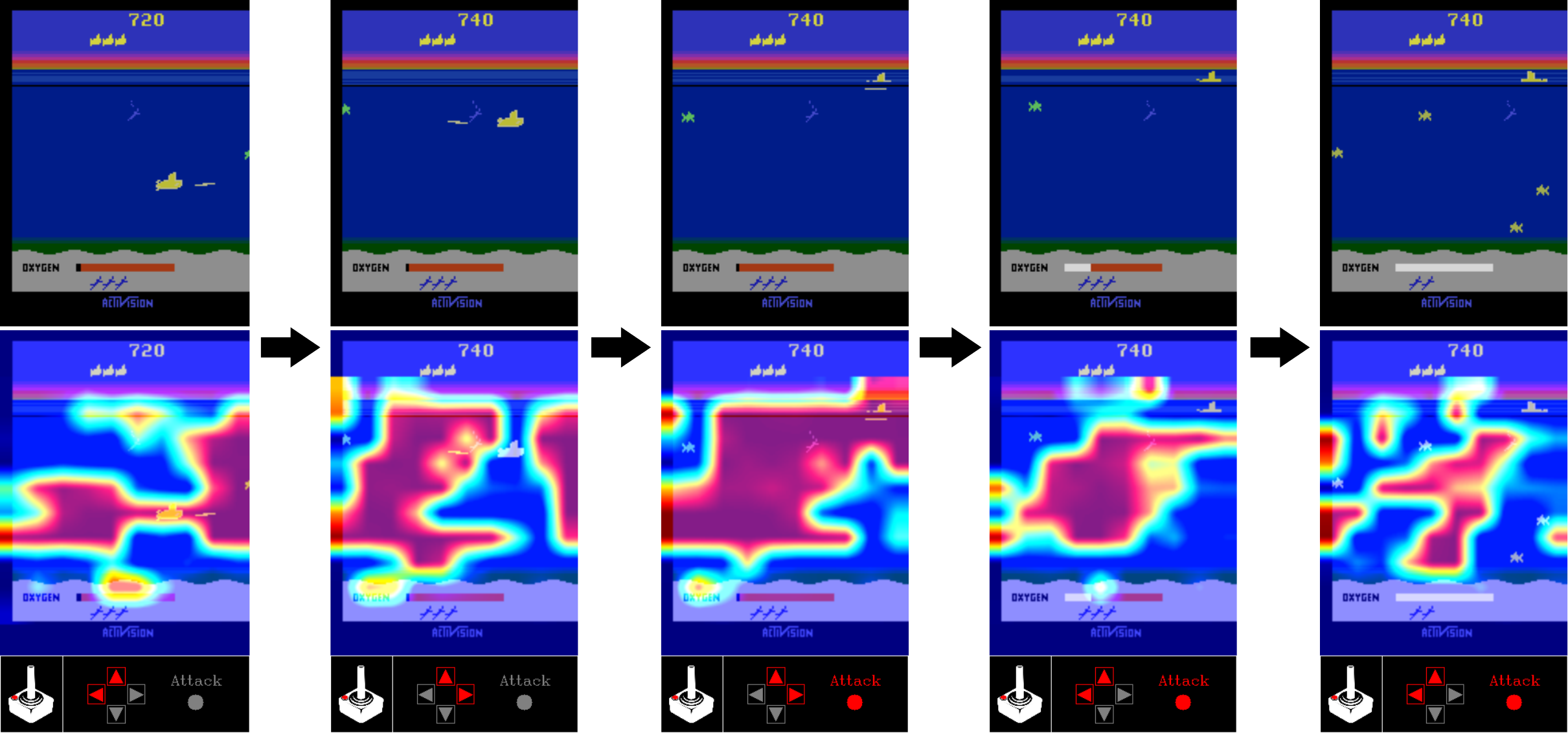}
	\caption{\textbf{Mask-attention in policy of Seaquest.}}
	\label{fig:appendix-policy-sq}
\end{figure*}

\begin{figure*}[h!]
	\centering
    \includegraphics[clip,width=\linewidth]{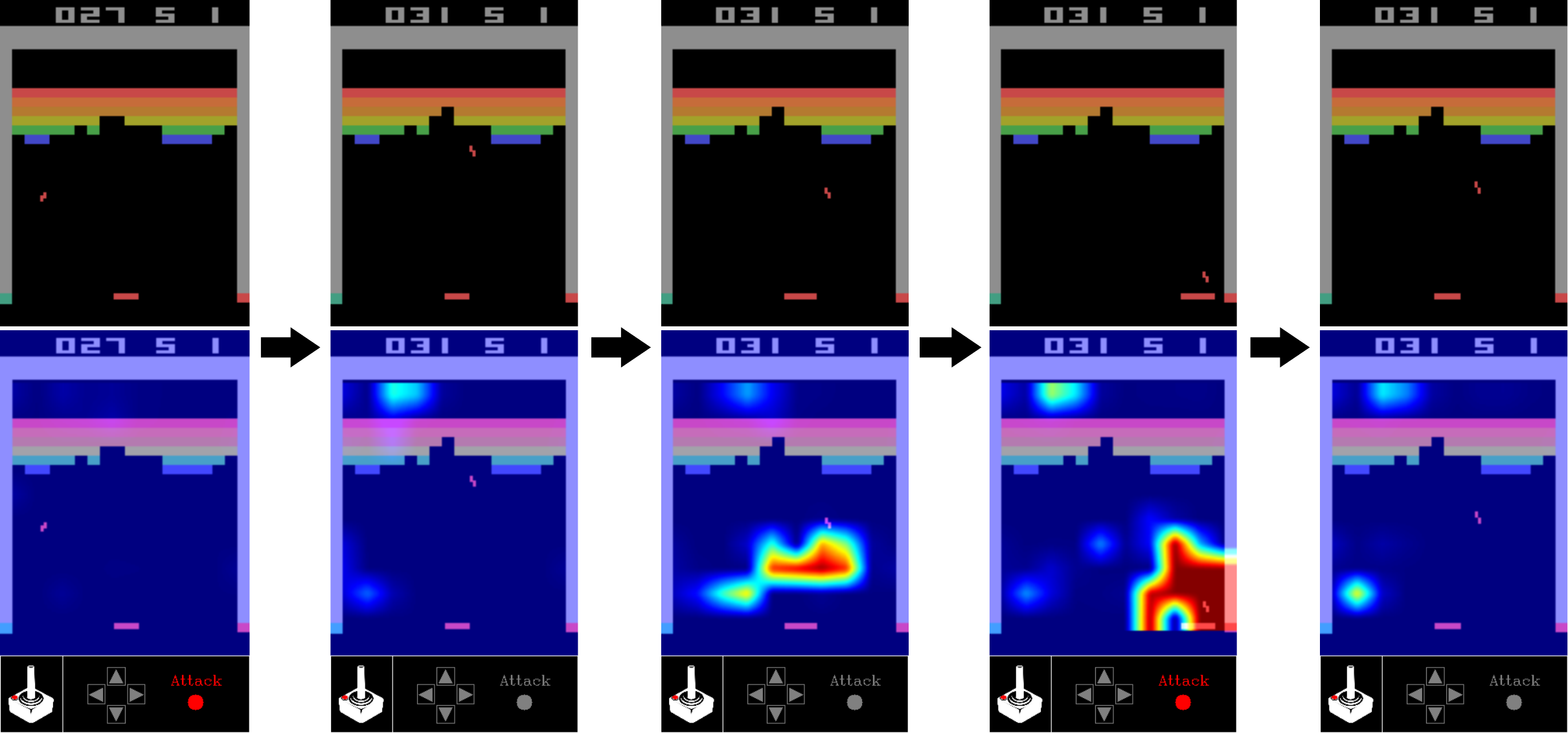}
	\caption{\textbf{Mask-attention in policy of Breakout.}}
	\label{fig:appendix-policy-bo}
\end{figure*}

\begin{figure*}[h!]
	\centering
    \includegraphics[clip,width=\linewidth]{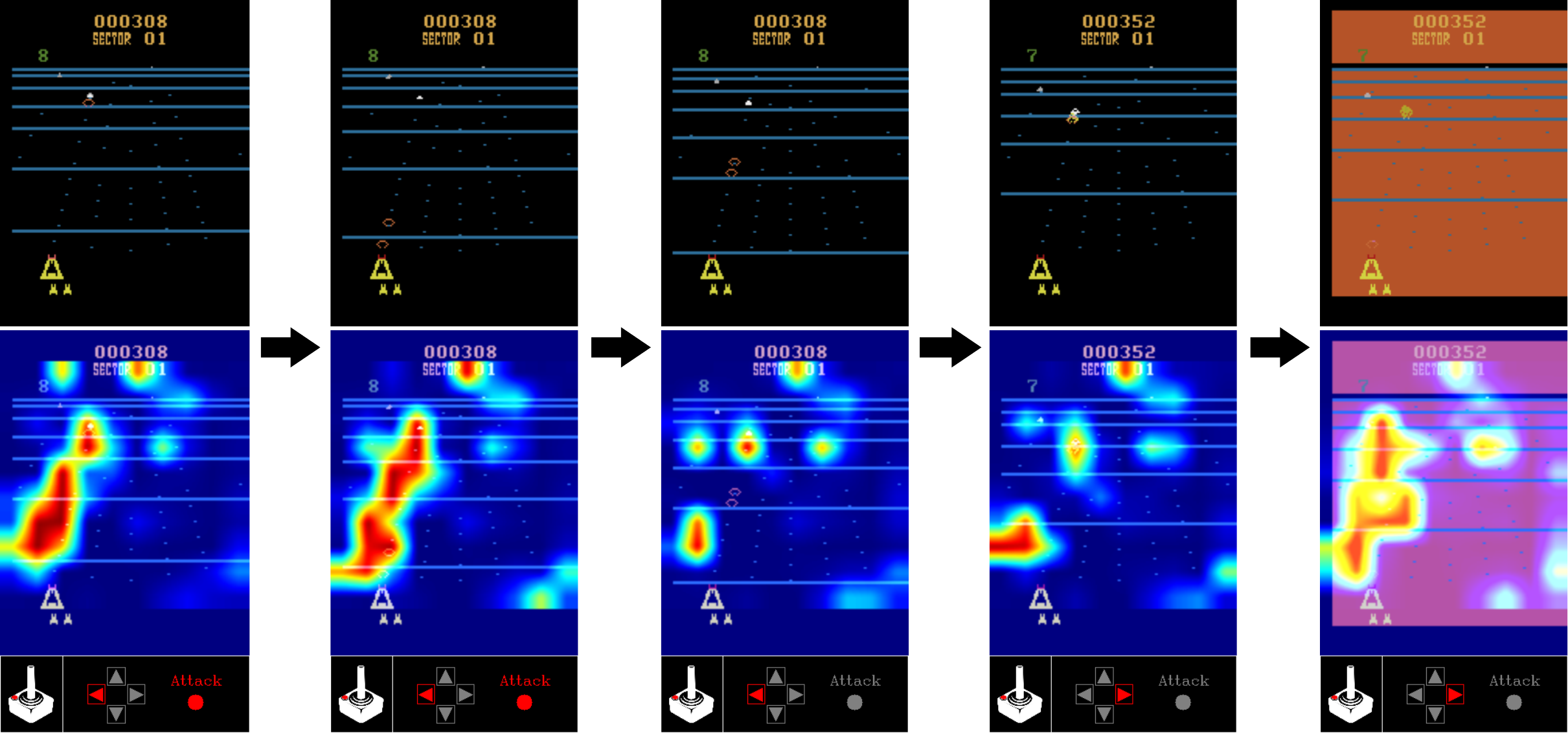}
	\caption{\textbf{Mask-attention in policy of Beamrider.}}
	\label{fig:appendix-policy-br}
\end{figure*}

\begin{figure*}[h!]
	\centering
    \includegraphics[clip,width=\linewidth]{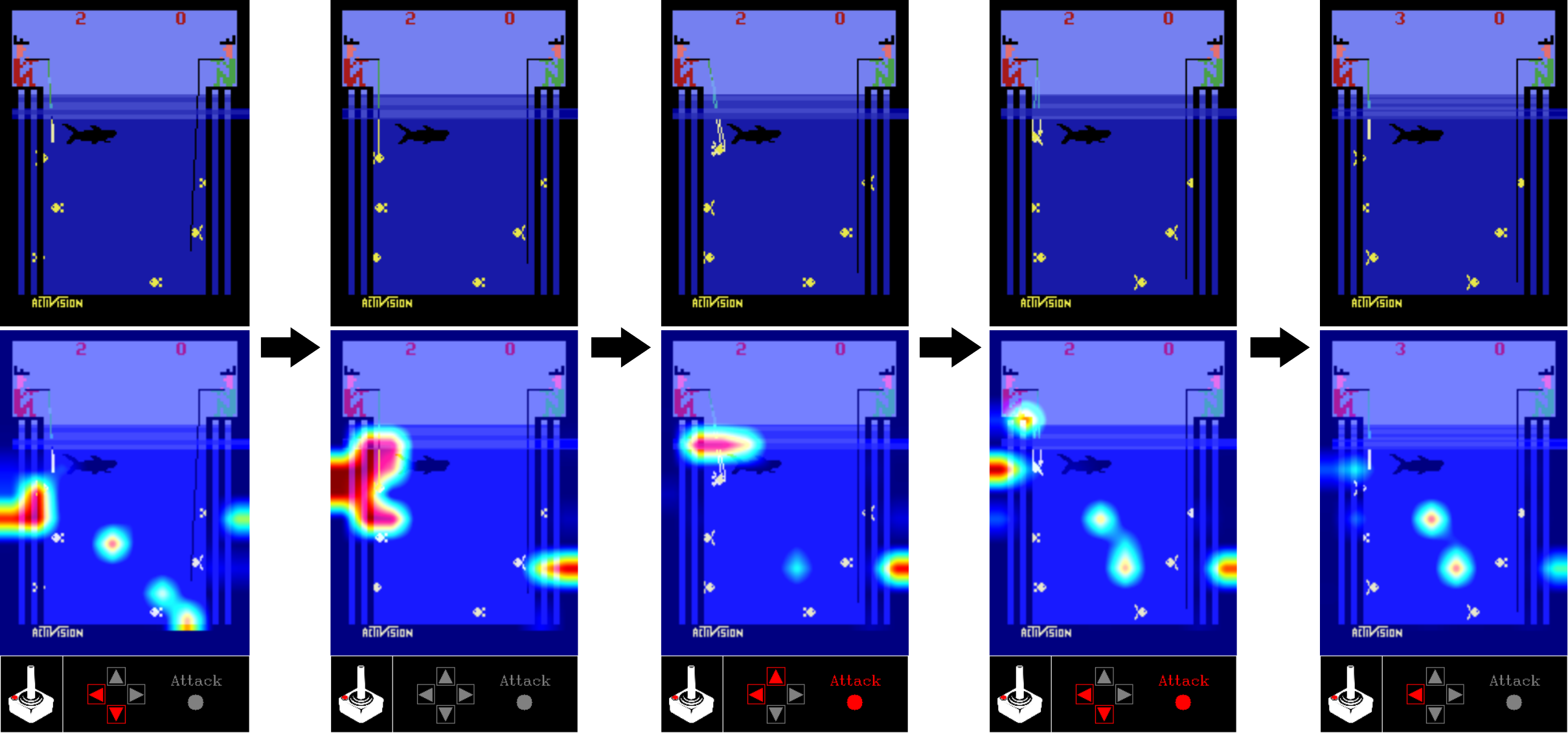}
	\caption{\textbf{Mask-attention in policy of Fishing Derby.}}
	\label{fig:appendix-policy-fd}
\end{figure*}

Figures \ref{fig:appendix-policy-ms}, \ref{fig:appendix-policy-si}, \ref{fig:appendix-policy-sq} \ref{fig:appendix-policy-bo}, \ref{fig:appendix-policy-br}, and \ref{fig:appendix-policy-fd} show mask-attentions in policy of each environment.

\clearpage

\subsubsection{\textbf{Mask-attentions in state value}}

\begin{figure*}[h!]
	\centering
    \includegraphics[clip,scale=0.64]{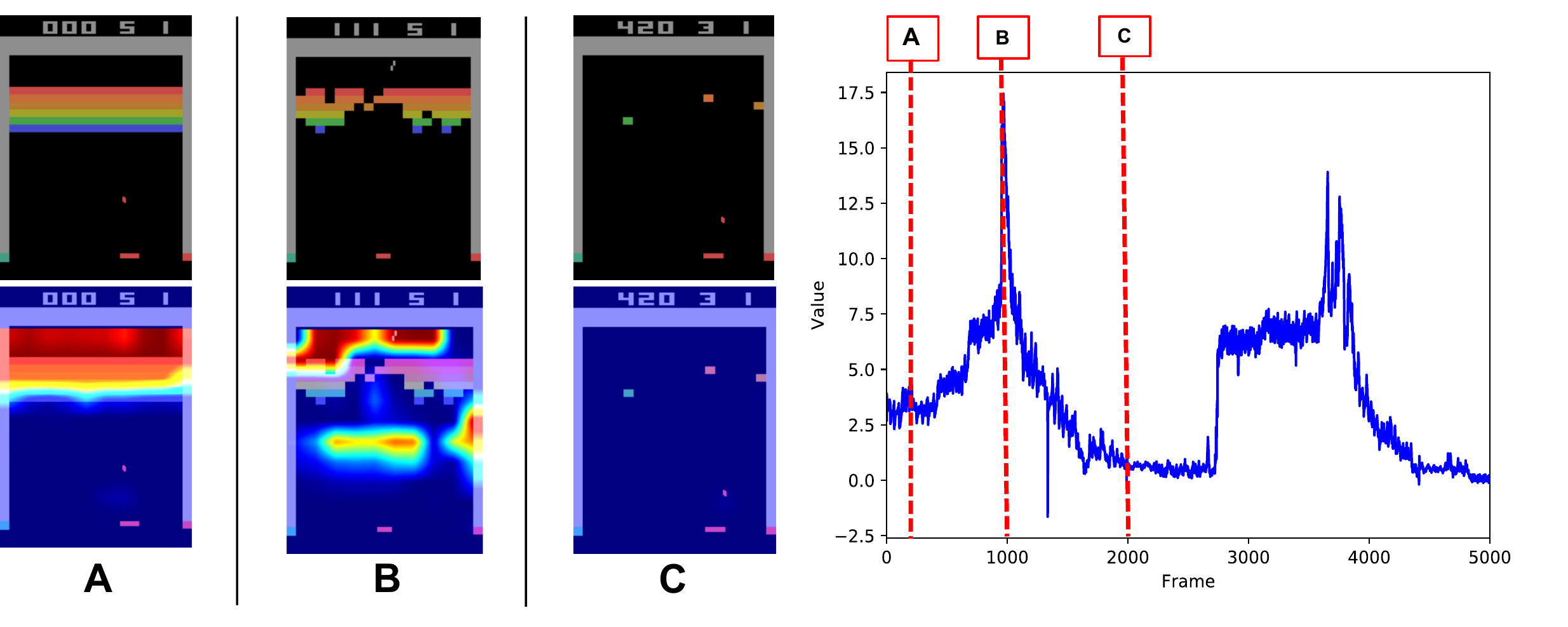}
	\caption{\textbf{Mask-attention in value of Breakout.}}
	\label{fig:appendix-value-bo}
\end{figure*}

\begin{figure*}[h!]
	\centering
    \includegraphics[clip,scale=0.64]{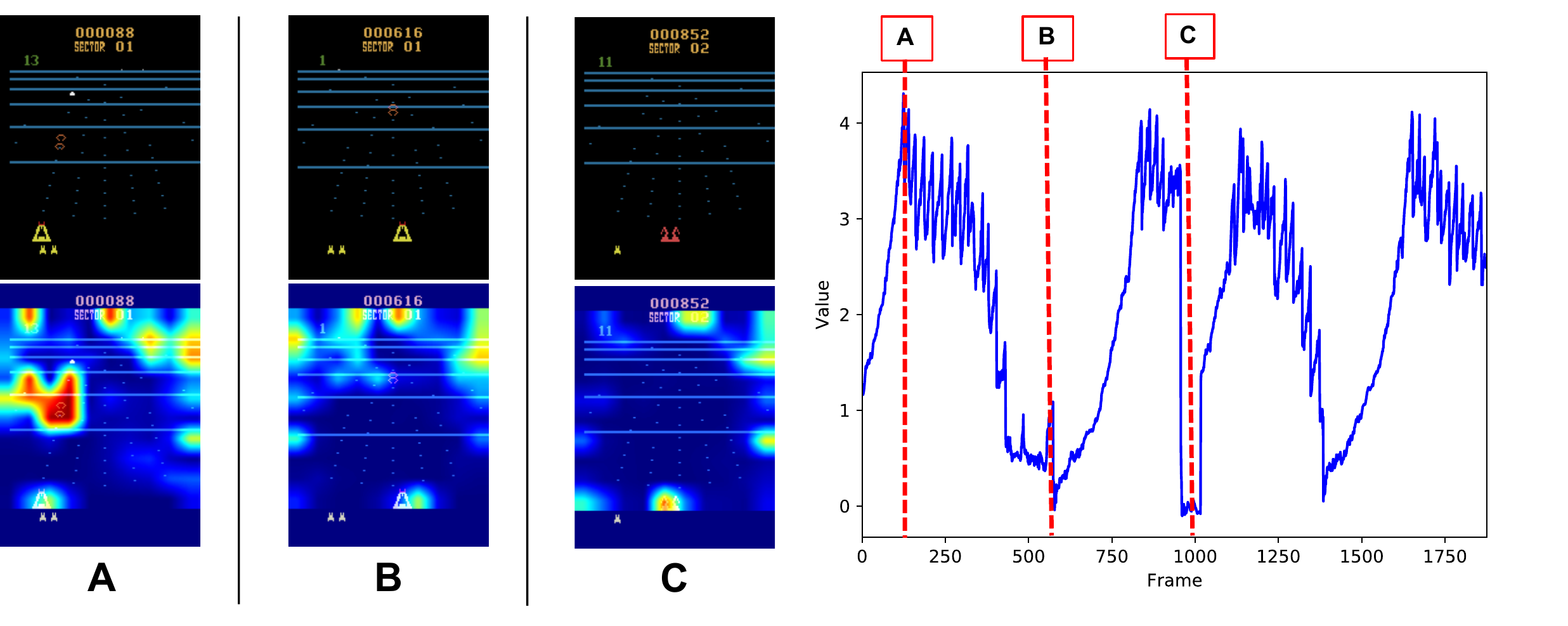}
	\caption{\textbf{Mask-attention in value of Beamrider.}}
	\label{fig:appendix-value-br}
\end{figure*}

\begin{figure*}[h!]
	\centering
    \includegraphics[clip,scale=0.64]{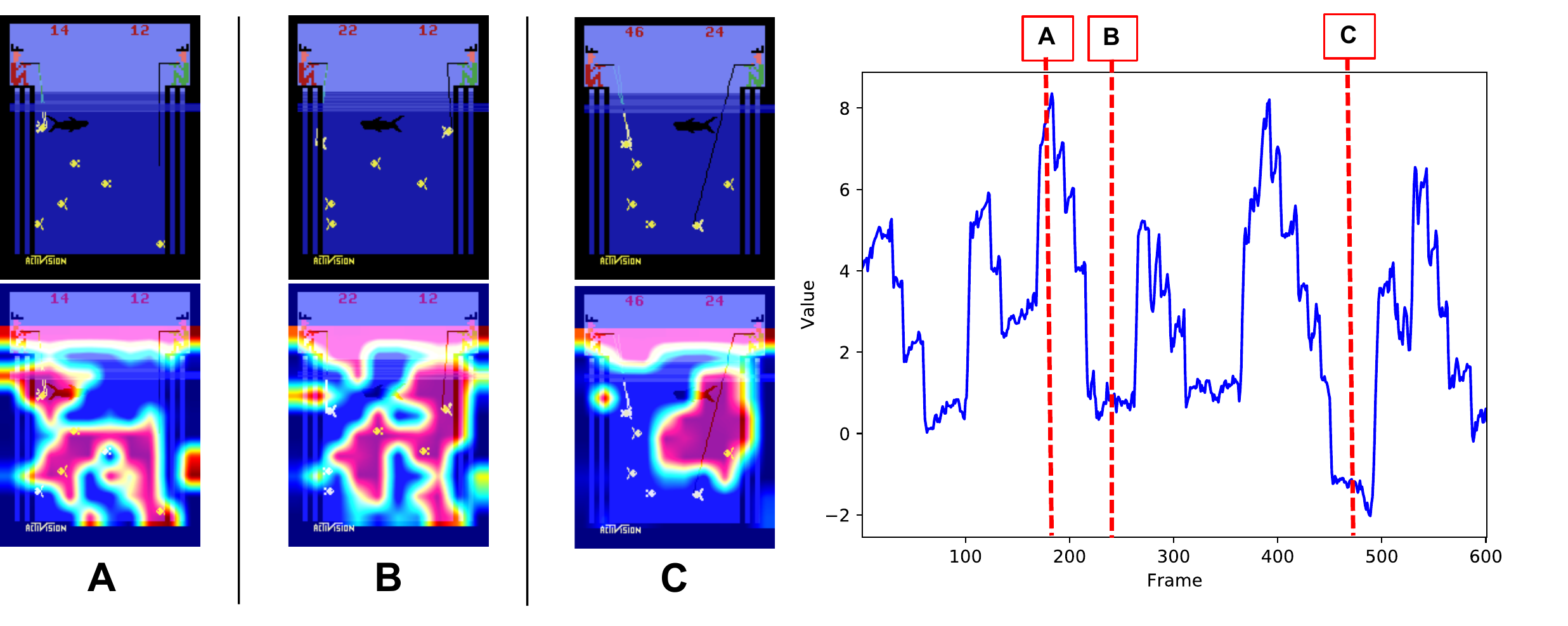}
	\caption{\textbf{Mask-attention in value of Fishing Derby.}}
	\label{fig:appendix-value-fd}
\end{figure*}

Figures \ref{fig:appendix-value-bo}, \ref{fig:appendix-value-br}, and \ref{fig:appendix-value-fd} show mask-attentions in state value.

\subsection{\textbf{Transitions of Mask-attentions in policy and state value}}

\begin{figure}[h!]
	\centering
    \includegraphics[clip,scale=0.82]{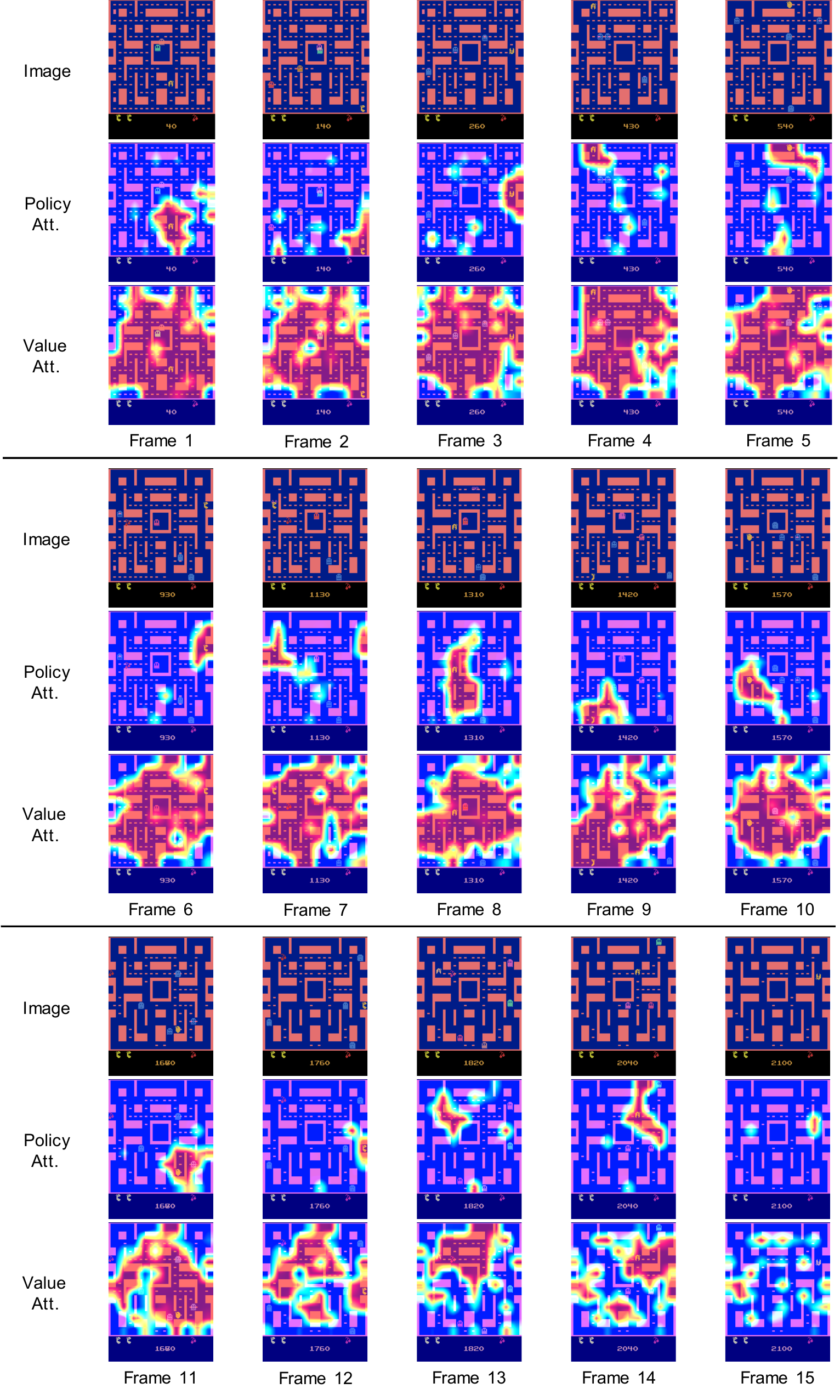}
	\caption{\textbf{Transitions of Mask-attentions in policy and value of Ms.Pac-Man}}
	\label{fig:appendix-policy-value-mp}
\end{figure}

\begin{figure}[h!]
	\centering
    \includegraphics[clip,scale=0.82]{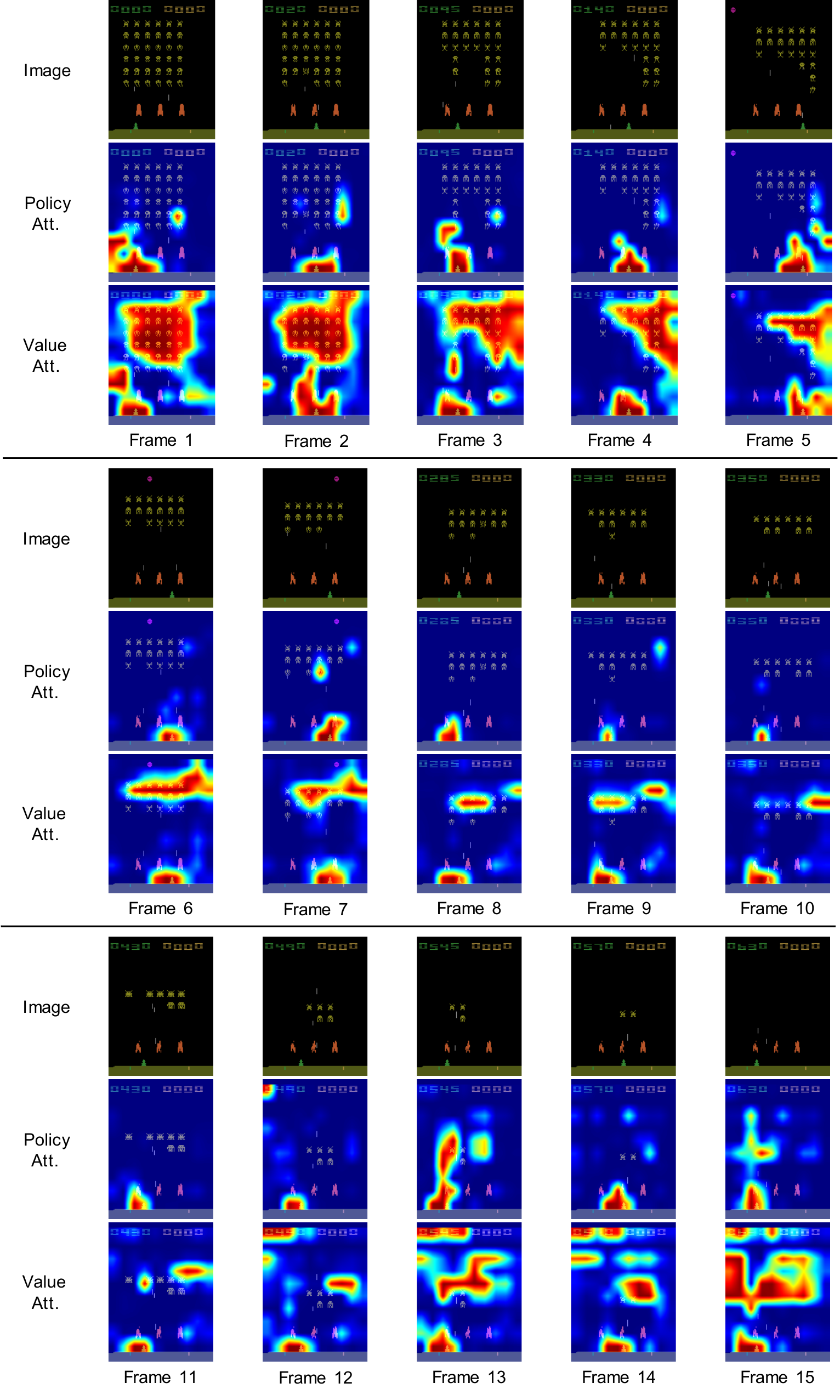}
	\caption{\textbf{Transitions of Mask-attentions in policy and value of Space Invaders}}
	\label{fig:appendix-policy-value-si}
\end{figure}

\begin{figure}[h!]
	\centering
    \includegraphics[clip,scale=0.82]{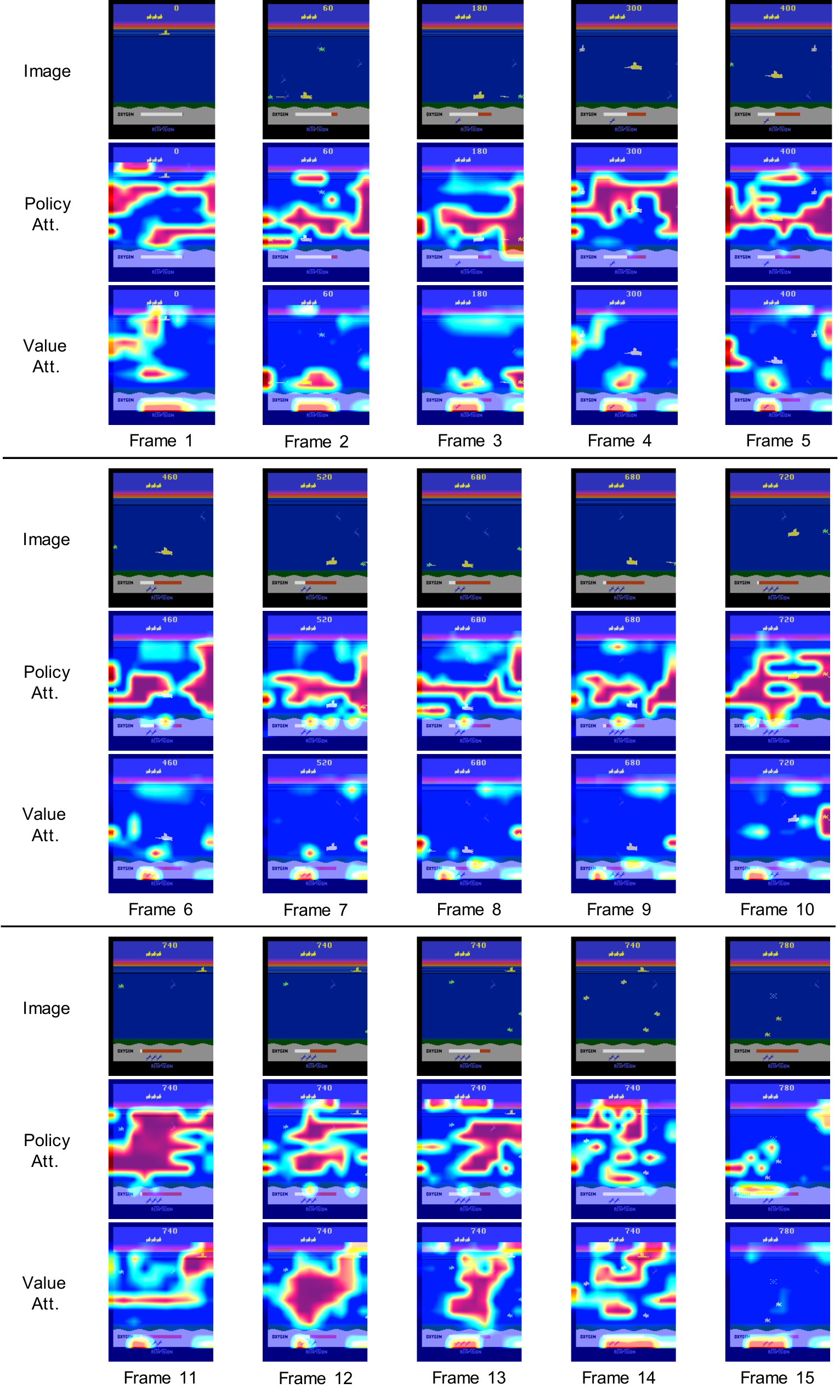}
	\caption{\textbf{Transitions of Mask-attentions in policy and value of Seaquest}}
	\label{fig:appendix-policy-value-sq}
\end{figure}

\begin{figure}[h!]
	\centering
    \includegraphics[clip,scale=0.82]{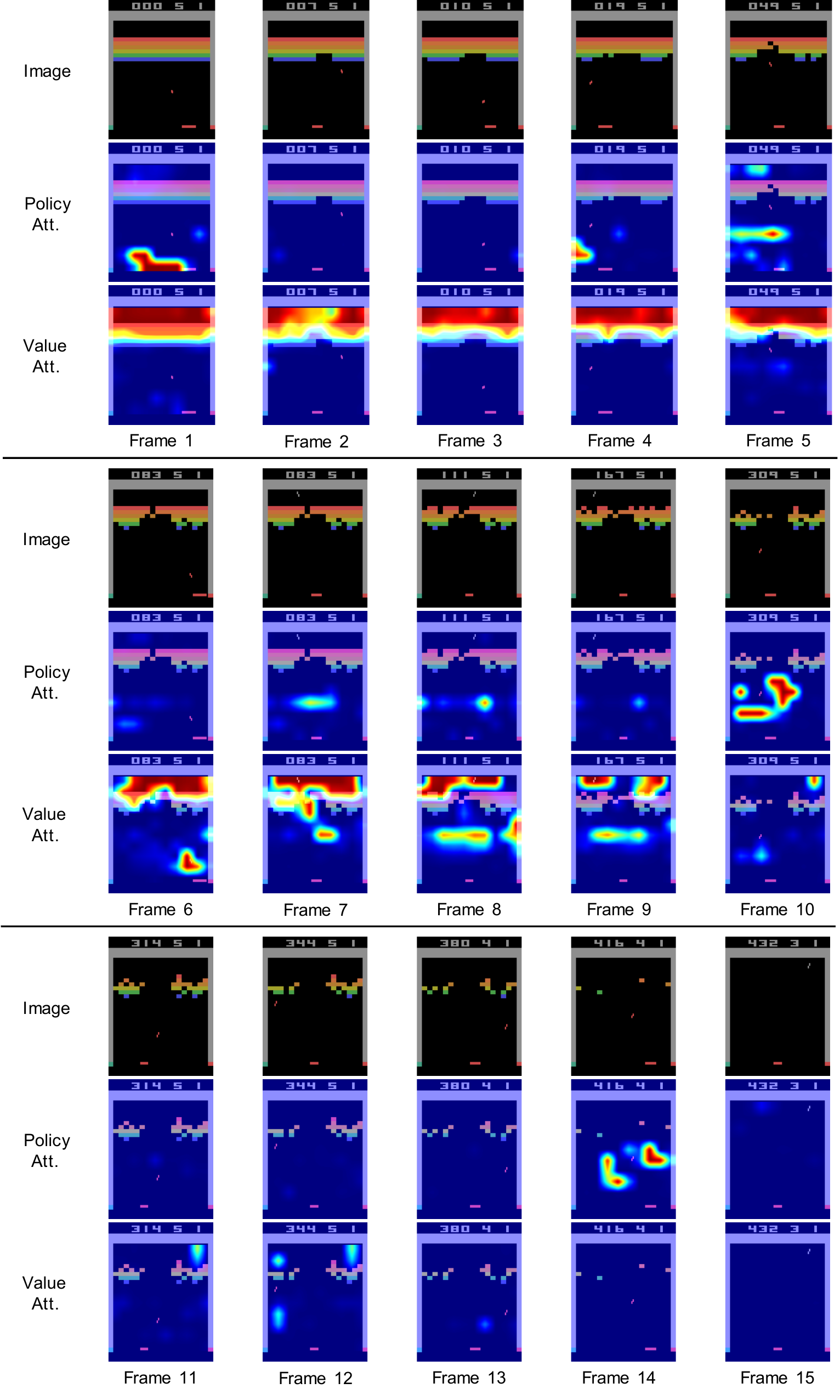}
	\caption{\textbf{Transitions of Mask-attentions in policy and value of Breakout}}
	\label{fig:appendix-policy-value-bo}
\end{figure}

\begin{figure}[h!]
	\centering
    \includegraphics[clip,scale=0.82]{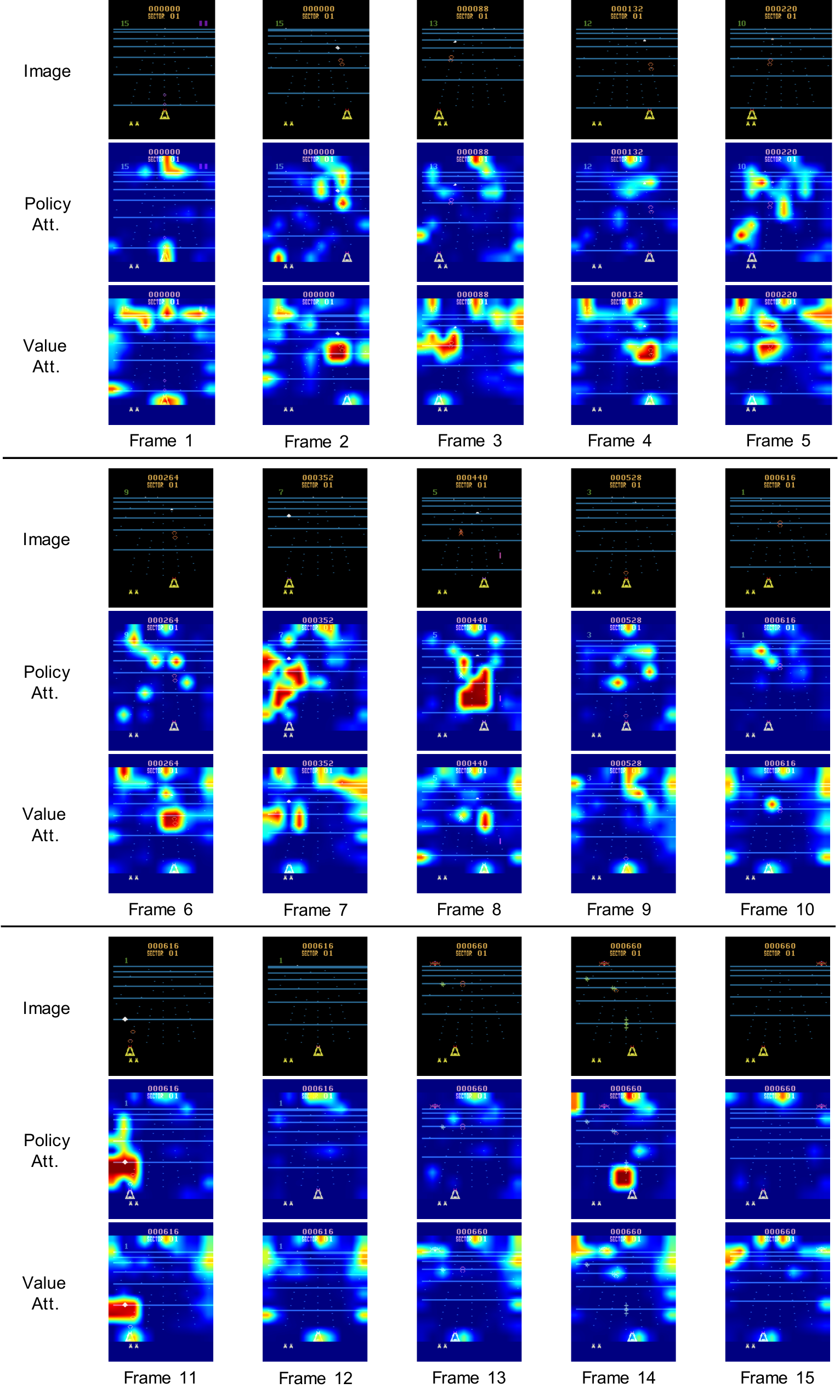}
	\caption{\textbf{Transitions of Mask-attentions in policy and value of Beamrider}}
	\label{fig:appendix-policy-value-br}
\end{figure}

\begin{figure}[h!]
	\centering
    \includegraphics[clip,scale=0.82]{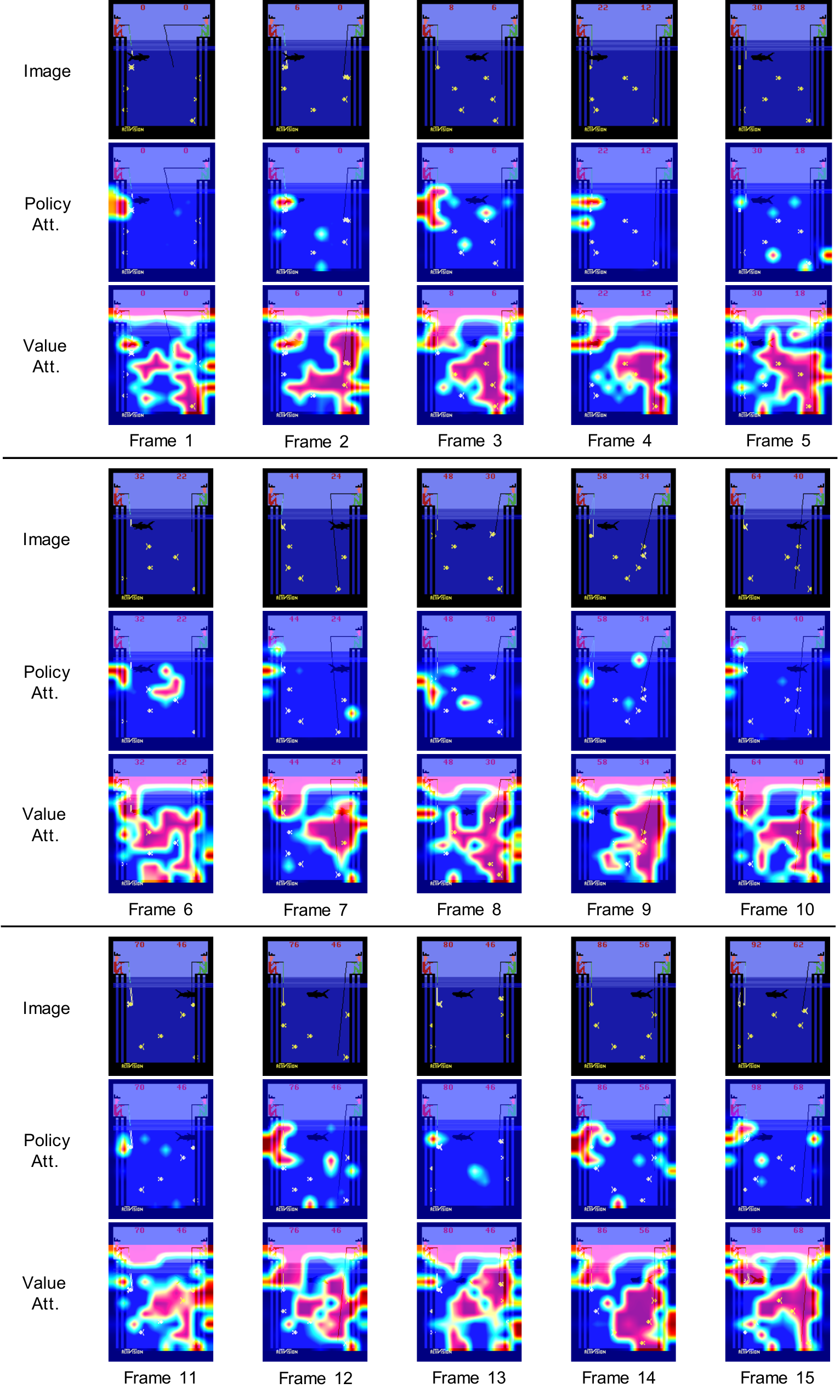}
	\caption{\textbf{Transitions of Mask-attentions in policy and value of Fishing Derby}}
	\label{fig:appendix-policy-value-fd}
\end{figure}

Figures \ref{fig:appendix-policy-value-mp}, \ref{fig:appendix-policy-value-si}, \ref{fig:appendix-policy-value-sq}, 
\ref{fig:appendix-policy-value-bo},
\ref{fig:appendix-policy-value-br}, and \ref{fig:appendix-policy-value-fd} show the transitions of mask-attentions in policy and state value.

\clearpage

\clearpage

\clearpage

\clearpage

\clearpage

\end{document}